\documentstyle[11pt]{article}
\newtheorem{definition}{Definition}[section]
\newtheorem{example}{Example}[section]
\newtheorem{remark}{Remark}[section]
\newtheorem{lemma}{Lemma}[section]
\newtheorem{theorem}[lemma]{Theorem}
\newtheorem{corollary}[lemma]{Corollary}

\begin{document}

\title{\Large \bf Linear Tabulated Resolution Based on Prolog
Control Strategy}

\author{Yi-Dong Shen\thanks{Work performed during a visit at
Department of Computing Science, 
University of Alberta, Canada.}\\
{\small  Department of Computer Science,
Chongqing University, Chongqing 400044, P.R.China}\\
{\small Email: ydshen@cs.ualberta.ca}\\[.1in]
Li-Yan Yuan  and Jia-Huai You\\
{\small  Department of Computing Science, University of
Alberta, Edmonton, Alberta, Canada T6G 2H1}\\
{\small  Email: \{yuan, you\}@cs.ualberta.ca}\\[.1in]
Neng-Fa Zhou \\
{\small Department of Computer and Information Science, Brooklyn College}\\ 
{\small The City University of New York, New York, NY 11210-2889, USA}\\
{\small Email: zhou@sci.brooklyn.cuny.edu}}

\date{}

\maketitle

\begin{abstract}
Infinite loops and redundant computations are long recognized
open problems in Prolog. 
Two ways have been explored to resolve these problems:
loop checking and tabling. Loop checking can cut infinite loops,
but it cannot be
both sound and complete even for function-free
logic programs. 
Tabling seems to be an effective way to resolve infinite loops and
redundant computations. However, existing tabulated
resolutions, such as OLDT-resolution, SLG-resolution, and Tabulated
SLS-resolution, are non-linear because they rely on
the solution-lookup mode in formulating tabling.
The principal disadvantage of non-linear resolutions is
that they cannot be implemented 
using a simple stack-based memory structure like that in Prolog.
Moreover, some strictly sequential operators such as cuts 
may not be handled as easily as in Prolog.

In this paper, we propose a hybrid method to resolve 
infinite loops and redundant computations.
We combine the ideas of loop checking and tabling to establish 
a linear tabulated resolution called
TP-resolution. TP-resolution has two
distinctive features: (1) It makes linear tabulated derivations
in the same way as Prolog except that
infinite loops are broken and redundant computations are reduced.
It handles cuts as effectively as Prolog.
(2) It is sound and complete for positive
logic programs with the bounded-term-size property.
The underlying algorithm can be implemented
by an extension to any existing Prolog abstract machines 
such as WAM or ATOAM.\\[.1in]
\end{abstract}

\section{Introduction}
While Prolog has many distinct advantages, 
it suffers from some serious problems,
among the best-known of which
are infinite loops and redundant computations. Infinite
loops cause users (especially less skilled users) to lose
confidence in writing terminating Prolog programs, whereas
redundant computations greatly reduce the
efficiency of Prolog. The existing approaches to 
resolving these problems can be classified into two categories:
loop checking and tabling.

Loop checking is a direct way to cut infinite loops. It
locates nodes at which SLD-derivations step into
a loop and prunes them from SLD-trees. Informally, an SLD-derivation
$G_0 \Rightarrow_{C_1,\theta_1}G_1 \Rightarrow ...$
$\Rightarrow_{C_i,\theta_i}G_i \Rightarrow ...$
$\Rightarrow_{C_k,\theta_k}G_k \Rightarrow ...$ is said to step
into a loop at a node $N_k$ labeled with a goal $G_k$ if there is a
node $N_i$ ($0 \leq i < k$) labeled with a goal $G_i$ in the
derivation such that $G_i$ and $G_k$ are {\em sufficiently
similar}. Many loop checking mechanisms
have been presented in the literature
(e.g. \cite{BAK91,Cov85,FPS95,shen97,Sk97,VG87,VL89}).
However, no loop checking mechanism can be 
both (weakly) sound and complete
because the loop checking problem itself is undecidable in
general even for function-free logic programs \cite{BAK91}.

The main idea of tabling is that during top-down query evaluation,
we store intermediate results of some subgoals 
and look them up to solve variants of
the subgoals that occur later.
Since no variant subgoals will be recomputed by applying the same
set of program clauses, infinite 
loops can be avoided. As a result, termination can be guaranteed
for bounded-term-size programs and redundant computations 
substantially reduced \cite{BD98,chen96,TS86,VL89,war92}.

There are many ways to formulate tabling, each leading to
a tabulated resolution (e.g. OLDT-resolution \cite{TS86}, SLG-resolution
\cite{chen96}, Tabulated SLS-resolution \cite{BD98}, etc.). However,
although existing tabulated resolutions differ in one aspect or
another, all of them rely on the so called {\em solution-lookup mode}.
That is, all nodes in a search tree/forest are
partitioned into two subsets, {\em solution} nodes and
{\em lookup} nodes; solution nodes produce 
child nodes using program clauses, whereas
lookup nodes produce child nodes using answers in tables.

Our investigation shows that 
the principal disadvantage of the solution-lookup mode
is that it makes tabulated resolutions non-linear. 
Let $G_0 \Rightarrow_{C_1, \theta_1} G_1 \Rightarrow$
$...  \Rightarrow_{C_i, \theta_i} G_i$ be the current
derivation with $G_i$ being the latest generated goal.
A tabulated resolution is said to be {\em linear}\footnote{
The concept of ``linear'' here is different from the one
used for SL-resolution \cite{KK71}.} if it 
makes the next derivation step either by expanding $G_i$
by resolving a subgoal in $G_i$ against
a program clause or a tabled answer, which yields 
$G_i \Rightarrow_{C_{i+1}, \theta_{i+1}} G_{i+1}$, 
or by expanding $G_{i-1}$ via backtracking. 
It is due to such non-linearity
that the underlying tabulated resolutions
cannot be implemented in the same way as SLD-resolution 
(Prolog) using a simple stack-based memory structure.
Moreover, some strictly sequential operators
such as cuts ($!$) may not be handled as easily as in Prolog.
For instance, in the well-known tabulated
resolution system XSB, clauses like

$\qquad p(.) \leftarrow ...,t(.),!,...$ 

\noindent where $t(.)$ is a tabled subgoal,
are not allowed because the tabled predicate $t$
occurs in the scope of a cut \cite{SSW94,SSWFR98}.

The objective of our research is to establish 
a hybrid approach to resolving 
infinite loops and redundant computations and develop a linear
tabulated Prolog system.
In this paper, we establish a theoretical framework for such a 
system, focusing on a linear tabulated resolution
$-$ {\em TP-resolution} for positive
logic programs (TP for Tabulated Prolog).

\begin{remark}
{\em
In this paper we will use the prefix {\em TP} to name
some key concepts such as TP-strategy, TP-tree, TP-derivation
and TP-resolution, in contrast to the standard Prolog control strategy,
Prolog-tree (i.e. SLD-tree generated under Prolog-strategy),
Prolog-derivation and Prolog-resolution (i.e. SLD-resolution
controlled by Prolog-strategy), respectively.
}
\end{remark}

In TP-resolution, each node in a search tree can act not only as a
solution node but also as a lookup node, regardless of when and where
it is generated. In fact, we do not distinguish between solution and lookup nodes 
in TP-resolution. This shows an essential difference from existing
tabulated resolutions using the solution-lookup mode.
The main idea is as follows: for
any selected tabled subgoal $A$ at a node $N_i$ 
labeled with a goal $G_i$, it
always first uses an answer $I$ in a table 
to generate a child node $N_{i+1}$ 
($N_i$ acts as a lookup node), which is labeled by 
the resolvent of $G_i$ and $I$; if no new answers
are available in the table, it resolves against
program clauses to produce child nodes 
($N_i$ then acts as a solution node).
The order in which answers in a table are used is based on
{\em first-generated-first-use}
and the order in which program 
clauses are applied is from top to bottom except for the case 
where the derivation steps into a loop at $N_i$. In such a case,
the subgoal $A$ skips the clause that is being
used by its closest ancestor subgoal that is a variant of $A$.
Like OLDT-resolution, TP-resolution is sound and complete for
positive logic programs with the bounded-term-size property.

The plan of this paper is as follows. In Section 2
we present a typical example to illustrate the main idea
of TP-resolution and its key differences from existing
tabulated resolutions. In Section 3, we formally define
TP-resolution. In Section 3.1 we discuss 
how to represent tables and how to 
operate on tables. In Section 3.2 we first introduce the so called
{\em PMF} mode for resolving tabled subgoals with program clauses,
which lays the basis for a linear tabulated resolution.
We then define a tabulated control strategy called {\em 
TP-strategy}, which enhances Prolog-strategy with proper policies for
the selection of answers in tables. Next we present a constructive
definition (an algorithm) of a {\em TP-tree} based on 
TP-strategy. Finally, based on TP-trees we define {\em
TP-derivations} and {\em TP-resolution}.

Section 4 is devoted to showing some major characteristics
of TP-resolution, including its termination property and
soundness and completeness. We also discuss in detail how
TP-resolution deals with the cut operator. 

We assume familiarity with the basic concepts of logic
programming, as presented in \cite{Ld87}. Here and throughout,
variables begin with a capital letter, and predicates, functions
and constants with a lower case letter.
By  $\vec{E}$ we denote a list/tuple ($E_1,...,E_m$) of elements.
Let $\vec{X}=(X_1,...,X_m)$ be a list of variables and
$\vec{I}=(I_1,...,I_m)$ a list of terms.
By $\vec{X}/\vec{I}$ we denote a substitution $\{X_1/I_1,...,X_m/I_m\}$.
By $p(.)$ we refer to any atom with
the predicate $p$ and by $p(\vec{X})$ to an atom $p(.)$
that contains the list $\vec{X}$ of distinct variables. For instance,
if $p(\vec{X})=p(W,a,f(Y),W)$, then $\vec{X}=(W,Y)$.
Let $G= \leftarrow A_1,...,A_m$
be a goal and $B$ a subgoal. By $G+B$ we denote the goal
$\leftarrow A_1,...,A_m,B$.
By a {\it variant} of an atom (resp. a subgoal or a
term) $A$ we mean an atom (resp. a subgoal or a term) $A'$ that is the same
as $A$ up to variable renaming.\footnote{By this definition,
$A$ is a variant of itself.} Let $V$ be a set of atoms
(resp. subgoals or terms) that are variants of each other;
then they are called {\em variant atoms} (resp. 
{\em variant subgoals} or {\em variant terms}).
Moreover, clauses with the same head predicate $p$ are
numbered sequentially, with $C_{p_i}$ referring to its
$i$-th clause $(i>0)$. Finally, unless otherwise
stated, by a (logic) program we refer to a positive logic
program with a finite set of clauses.

\section{An Illustrative Example}

We use the following simple program to illustrate
the basic idea of the TP approach. For convenience of
presentation, we choose OLDT-resolution \cite{TS86} for a
side-by-side comparison (other typical tabulated resolutions,
such as SLG-resolution
\cite{chen96} and Tabulated SLS-resolution \cite{BD98}, have similar
effects).
\begin{tabbing}
\hspace{.2in} $P_1$: \= $reach(X,Y) \leftarrow reach(X,Z),edge(Z,Y).$
\`$C_{r_1}$ \\
\> $reach(X,X).$ \`$C_{r_2}$ \\
\> $reach(X,d).$ \`$C_{r_3}$ \\[.1in]
\> $edge(a,b).$ \`$C_{e_1}$ \\
\> $edge(d,e).$ \`$C_{e_2}$
\end{tabbing}

Let $G_0 = \leftarrow reach(a,X)$ be the query (top goal). Then Prolog
will step into an infinite
loop right after the application of the first clause
$C_{r_1}$. We now show how it works using OLDT-resolution
(under the depth-first control strategy). Starting
from the root node $N_0$ labeled with the goal $\leftarrow reach(a,X)$,
the application of the clause $C_{r_1}$ gives a child node $N_1$
labeled with the goal $\leftarrow reach(a,Z),edge(Z,X)$ (see Figure \ref{fig1}).
Since the subgoal $reach(a,Z)$ is a variant of $reach(a,X)$ that
occurred earlier, it is suspended to wait for $reach(a,X)$ to 
produce answers. $N_0$ and $N_1$ (resp. $reach(a,X)$ and $reach(a,Z)$)
are then called {\em
solution} and {\em lookup} nodes (resp. subgoals),
respectively. So the derivation
goes back to $N_0$ and resolves $reach(a,X)$ with the second
clause $C_{r_2}$,
which gives a sibling node $N_2$ labeled with an empty
clause $\Box$.
Since  $reach(a,a)$ is an answer to the subgoal $reach(a,X)$,
it is memorized in a table, say $TB(reach(a,X))$. The derivation
then jumps back to $N_1$ and uses the answer $reach(a,a)$ in the
table to resolve with the lookup subgoal $reach(a,Z)$,
which gives a new node $N_3$ labeled with $\leftarrow edge(a,X)$.
Next, the node $N_4$ labeled with $\Box$
is derived from $N_3$ by resolving the subgoal $edge(a,X)$
with the clause $C_{e_1}$. So the answer $reach(a,b)$ is added
to the table $TB(reach(a,X))$. After these steps, the
OLDT-derivation evolves into a tree as depicted in Figure \ref{fig1},
which is clearly not linear.

\begin{figure}[h]
\setlength{\unitlength}{3947sp}%
\begingroup\makeatletter\ifx\SetFigFont\undefined%
\gdef\SetFigFont#1#2#3#4#5{%
  \reset@font\fontsize{#1}{#2pt}%
  \fontfamily{#3}\fontseries{#4}\fontshape{#5}%
  \selectfont}%
\fi\endgroup%
\begin{picture}(4200,1917)(2626,-2857)
\thicklines
\put(4876,-2311){\vector( 0,-1){300}}
\put(4876,-1711){\vector( 0,-1){300}}
\put(4876,-1111){\vector( 0,-1){300}}
\put(5633,-1163){\vector( 3,-1){675}}
\put(4951,-2536){\makebox(0,0)[lb]{\smash{\SetFigFont{8}{9.6}{\rmdefault}{\mddefault}{\updefault}$C_{e_1}$}}}
\put(4951,-1861){\makebox(0,0)[lb]{\smash{\SetFigFont{8}{9.6}{\rmdefault}{\mddefault}{\updefault}$Z=a$ (Get $reach(a,a)$ from the table)}}}
\put(4951,-1261){\makebox(0,0)[lb]{\smash{\SetFigFont{8}{9.6}{\rmdefault}{\mddefault}{\updefault}$C_{r_1}$}}}
\put(6076,-1261){\makebox(0,0)[lb]{\smash{\SetFigFont{8}{9.6}{\rmdefault}{\mddefault}{\updefault}$C_{r_2}$}}}
\put(5026,-2836){\makebox(0,0)[lb]{\smash{\SetFigFont{8}{9.6}{\rmdefault}{\mddefault}{\updefault}(Add $reach(a,b)$ to the table)}}}
\put(4501,-2836){\makebox(0,0)[lb]{\smash{\SetFigFont{9}{10.8}{\rmdefault}{\mddefault}{\updefault}$N_4$:  $\Box$}}}
\put(6826,-1636){\makebox(0,0)[lb]{\smash{\SetFigFont{8}{9.6}{\rmdefault}{\mddefault}{\updefault}(Add $reach(a,a)$ to the table)}}}
\put(2926,-1036){\makebox(0,0)[lb]{\smash{\SetFigFont{9}{10.8}{\rmdefault}{\mddefault}{\updefault}$\qquad\qquad\qquad\qquad$ $N_0$:  $\leftarrow reach(a, X)$}}}
\put(2626,-1636){\makebox(0,0)[lb]{\smash{\SetFigFont{9}{10.8}{\rmdefault}{\mddefault}{\updefault}$\qquad\qquad\qquad\qquad$ $N_1$:  $\leftarrow reach(a, Z),edge(Z,X)$$\qquad$ $N_2$: $\Box$ }}}
\put(2926,-2236){\makebox(0,0)[lb]{\smash{\SetFigFont{9}{10.8}{\rmdefault}{\mddefault}{\updefault}$\qquad\qquad\qquad\qquad$ $N_3$:  $\leftarrow edge(a, X)$}}}
\end{picture}

\caption{OLDT-derivation.$\qquad\qquad\qquad\qquad\qquad\qquad$}\label{fig1}
\end{figure}

We now explain how TP-resolution works. Starting from the root
node $N_0$ labeled with the goal $\leftarrow reach(a,X)$
we apply the clause $C_{r_1}$ to derive a child node $N_1$
labeled with the goal $\leftarrow reach(a,Z),edge(Z,X)$ (see Figure \ref{fig2}).
As the subgoal $reach(a,Z)$ is a variant of $reach(a,X)$
and the latter is an {\em ancestor} of the former (i.e.,
the derivation steps into a loop at $N_1$ \cite{shen97}), we choose
$C_{r_2}$, the clause from the backtracking point of
the subgoal $reach(a,X)$, to
resolve with $reach(a,Z)$, which gives a child node $N_2$
labeled with $\leftarrow edge(a,X)$. Since $reach(a,a)$ is an
answer to the subgoal $reach(a,Z)$, it
is memorized in a table $TB(reach(a,X))$.
We then resolve the subgoal $edge(a,X)$ against
the clause $C_{e_1}$, which gives the leaf
$N_3$ labeled with $\Box$. So the answer $reach(a,b)$ to
the subgoal $reach(a,X)$ is
added to the table $TB(reach(a,X))$. After these steps,
we get a path as shown in Figure \ref{fig2}, which is clearly linear.

\begin{figure}[h]
\setlength{\unitlength}{3947sp}%
\begingroup\makeatletter\ifx\SetFigFont\undefined%
\gdef\SetFigFont#1#2#3#4#5{%
  \reset@font\fontsize{#1}{#2pt}%
  \fontfamily{#3}\fontseries{#4}\fontshape{#5}%
  \selectfont}%
\fi\endgroup%
\begin{picture}(2400,1917)(2626,-2857)
\thicklines
\put(4876,-2311){\vector( 0,-1){300}}
\put(4876,-1711){\vector( 0,-1){300}}
\put(4876,-1111){\vector( 0,-1){300}}
\put(4951,-2536){\makebox(0,0)[lb]{\smash{\SetFigFont{8}{9.6}{\rmdefault}{\mddefault}{\updefault}$C_{e_1}$}}}
\put(4951,-1861){\makebox(0,0)[lb]{\smash{\SetFigFont{8}{9.6}{\rmdefault}{\mddefault}{\updefault}$C_{r_2}$ (Add $reach(a,a)$ to the table)}}}
\put(4951,-1261){\makebox(0,0)[lb]{\smash{\SetFigFont{8}{9.6}{\rmdefault}{\mddefault}{\updefault}$C_{r_1}$}}}
\put(5026,-2836){\makebox(0,0)[lb]{\smash{\SetFigFont{8}{9.6}{\rmdefault}{\mddefault}{\updefault}(Add $reach(a,b)$ to the table)}}}
\put(4501,-2836){\makebox(0,0)[lb]{\smash{\SetFigFont{9}{10.8}{\rmdefault}{\mddefault}{\updefault}$N_3$:  $\Box$}}}
\put(2926,-1036){\makebox(0,0)[lb]{\smash{\SetFigFont{9}{10.8}{\rmdefault}{\mddefault}{\updefault}$\qquad\qquad\qquad\qquad$ $N_0$:  $\leftarrow reach(a, X)$}}}
\put(2626,-1636){\makebox(0,0)[lb]{\smash{\SetFigFont{9}{10.8}{\rmdefault}{\mddefault}{\updefault}$\qquad\qquad\qquad\qquad$ $N_1$:  $\leftarrow reach(a, Z),edge(Z,X)$}}}
\put(2926,-2236){\makebox(0,0)[lb]{\smash{\SetFigFont{9}{10.8}{\rmdefault}{\mddefault}{\updefault}$\qquad\qquad\qquad\qquad$ $N_2$:  $\leftarrow edge(a, X)$}}}
\end{picture}

\caption{TP-derivation.$\qquad\qquad\qquad\qquad\qquad\qquad$}\label{fig2}
\end{figure}

Now consider backtracking. Remember that after the above derivation
steps, the table $TB($ $reach(a,X))$ consists of two
answers, $reach(a,a)$ and $reach(a,b)$.
For the OLDT approach, it first backtracks
to $N_3$ and then to $N_1$ (Figure \ref{fig1}). Since the subgoal
$reach(a,Z)$ has used the first answer in the table before, it resolves with
the second, $reach(a,b)$, which gives a new node
labeled with the goal $\leftarrow edge(b,X)$. Obviously, this
goal will fail, so it backtracks to $N_1$ again.
This time no new answers in the table are available to
the subgoal $reach(a,Z)$, so it is suspended and the derivation
goes to the solution node
$N_0$. The third clause $C_{r_3}$ is then selected to resolve
with the subgoal $reach(a,X)$, yielding a new answer $reach(a,d)$,
which is added to the table. The derivation then goes back
to $N_1$ where the new answer is used in the same way as
described before.

\begin{figure}[h]
\centering
\setlength{\unitlength}{3947sp}%
\begingroup\makeatletter\ifx\SetFigFont\undefined%
\gdef\SetFigFont#1#2#3#4#5{%
  \reset@font\fontsize{#1}{#2pt}%
  \fontfamily{#3}\fontseries{#4}\fontshape{#5}%
  \selectfont}%
\fi\endgroup%
\begin{picture}(6537,1905)(901,-2857)
\thicklines
\put(1951,-2311){\vector( 0,-1){300}}
\put(4951,-2311){\vector( 0,-1){300}}
\put(4276,-1111){\vector( 0,-1){300}}
\put(3358,-1640){\vector(-4,-1){1482.353}}
\put(4351,-1711){\vector( 3,-2){450}}
\put(4210,-1723){\vector(-4,-3){384}}
\put(5591,-1627){\vector( 4,-1){1835.294}}
\put(2026,-2461){\makebox(0,0)[lb]{\smash{\SetFigFont{8}{9.6}{\rmdefault}{\mddefault}{\updefault}$C_{e_1}$}}}
\put(5026,-2461){\makebox(0,0)[lb]{\smash{\SetFigFont{8}{9.6}{\rmdefault}{\mddefault}{\updefault}$C_{e_2}$}}}
\put(3601,-1036){\makebox(0,0)[lb]{\smash{\SetFigFont{9}{10.8}{\rmdefault}{\mddefault}{\updefault}$N_0$:  $\leftarrow reach(a, X)$}}}
\put(4351,-1261){\makebox(0,0)[lb]{\smash{\SetFigFont{8}{9.6}{\rmdefault}{\mddefault}{\updefault}$C_{r_1}$}}}
\put(5101,-2836){\makebox(0,0)[lb]{\smash{\SetFigFont{8}{9.6}{\rmdefault}{\mddefault}{\updefault}(Add $reach(a,e)$)}}}
\put(3376,-1561){\makebox(0,0)[lb]{\smash{\SetFigFont{9}{10.8}{\rmdefault}{\mddefault}{\updefault}$N_1$:  $\leftarrow reach(a, Z),edge(Z,X)$ }}}
\put(901,-1861){\makebox(0,0)[lb]{\smash{\SetFigFont{8}{9.6}{\rmdefault}{\mddefault}{\updefault}$C_{r_2}$ (Add $reach(a,a)$)}}}
\put(6826,-2236){\makebox(0,0)[lb]{\smash{\SetFigFont{9}{10.8}{\rmdefault}{\mddefault}{\updefault}$N_7$:  $\leftarrow edge(e, X)$}}}
\put(2851,-1936){\makebox(0,0)[lb]{\smash{\SetFigFont{8}{9.6}{\rmdefault}{\mddefault}{\updefault}Get $reach(a,b)$}}}
\put(4801,-1936){\makebox(0,0)[lb]{\smash{\SetFigFont{8}{9.6}{\rmdefault}{\mddefault}{\updefault}$C_{r_3}$ (Add $reach(a,d)$)}}}
\put(4501,-2236){\makebox(0,0)[lb]{\smash{\SetFigFont{9}{10.8}{\rmdefault}{\mddefault}{\updefault}$N_5$:  $\leftarrow edge(d, X)$}}}
\put(3001,-2236){\makebox(0,0)[lb]{\smash{\SetFigFont{9}{10.8}{\rmdefault}{\mddefault}{\updefault}$N_4$:  $\leftarrow edge(b, X)$}}}
\put(7051,-1936){\makebox(0,0)[lb]{\smash{\SetFigFont{8}{9.6}{\rmdefault}{\mddefault}{\updefault}Get $reach(a,e)$}}}
\put(1276,-2161){\makebox(0,0)[lb]{\smash{\SetFigFont{9}{10.8}{\rmdefault}{\mddefault}{\updefault}$N_2$:  $\leftarrow edge(a, X)$}}}
\put(2101,-2836){\makebox(0,0)[lb]{\smash{\SetFigFont{8}{9.6}{\rmdefault}{\mddefault}{\updefault}(Add $reach(a,b)$)}}}
\put(4576,-2836){\makebox(0,0)[lb]{\smash{\SetFigFont{9}{10.8}{\rmdefault}{\mddefault}{\updefault}$N_6$:  $\Box$}}}
\put(1651,-2836){\makebox(0,0)[lb]{\smash{\SetFigFont{9}{10.8}{\rmdefault}{\mddefault}{\updefault}$N_3$:  $\Box$}}}
\end{picture}

\caption{TP-derivations of $P_1 \cup \{G_0\}$.}\label{fig3}
\end{figure}
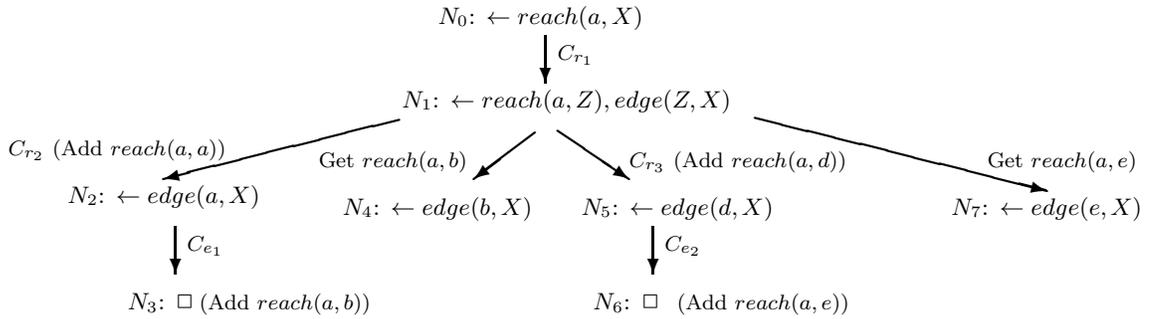

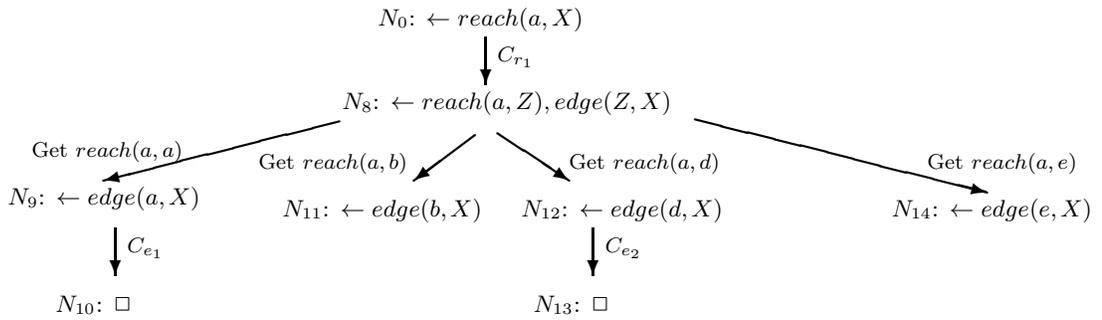
\begin{figure}[h]
\centering
\setlength{\unitlength}{3947sp}%
\begingroup\makeatletter\ifx\SetFigFont\undefined%
\gdef\SetFigFont#1#2#3#4#5{%
  \reset@font\fontsize{#1}{#2pt}%
  \fontfamily{#3}\fontseries{#4}\fontshape{#5}%
  \selectfont}%
\fi\endgroup%
\begin{picture}(6162,1905)(1276,-2857)
\thicklines
\put(1951,-2311){\vector( 0,-1){300}}
\put(4951,-2311){\vector( 0,-1){300}}
\put(4276,-1111){\vector( 0,-1){300}}
\put(3358,-1640){\vector(-4,-1){1482.353}}
\put(4351,-1711){\vector( 3,-2){450}}
\put(4210,-1723){\vector(-4,-3){384}}
\put(5591,-1627){\vector( 4,-1){1835.294}}
\put(2026,-2461){\makebox(0,0)[lb]{\smash{\SetFigFont{8}{9.6}{\rmdefault}{\mddefault}{\updefault}$C_{e_1}$}}}
\put(5026,-2461){\makebox(0,0)[lb]{\smash{\SetFigFont{8}{9.6}{\rmdefault}{\mddefault}{\updefault}$C_{e_2}$}}}
\put(3601,-1036){\makebox(0,0)[lb]{\smash{\SetFigFont{9}{10.8}{\rmdefault}{\mddefault}{\updefault}$N_0$:  $\leftarrow reach(a, X)$}}}
\put(4351,-1261){\makebox(0,0)[lb]{\smash{\SetFigFont{8}{9.6}{\rmdefault}{\mddefault}{\updefault}$C_{r_1}$}}}
\put(3376,-1561){\makebox(0,0)[lb]{\smash{\SetFigFont{9}{10.8}{\rmdefault}{\mddefault}{\updefault}$N_8$:  $\leftarrow reach(a, Z),edge(Z,X)$ }}}
\put(6826,-2236){\makebox(0,0)[lb]{\smash{\SetFigFont{9}{10.8}{\rmdefault}{\mddefault}{\updefault}$N_{14}$:  $\leftarrow edge(e, X)$}}}
\put(2851,-1936){\makebox(0,0)[lb]{\smash{\SetFigFont{8}{9.6}{\rmdefault}{\mddefault}{\updefault}Get $reach(a,b)$}}}
\put(4801,-1936){\makebox(0,0)[lb]{\smash{\SetFigFont{8}{9.6}{\rmdefault}{\mddefault}{\updefault}Get $reach(a,d)$}}}
\put(4501,-2236){\makebox(0,0)[lb]{\smash{\SetFigFont{9}{10.8}{\rmdefault}{\mddefault}{\updefault}$N_{12}$:  $\leftarrow edge(d, X)$}}}
\put(3001,-2236){\makebox(0,0)[lb]{\smash{\SetFigFont{9}{10.8}{\rmdefault}{\mddefault}{\updefault}$N_{11}$:  $\leftarrow edge(b, X)$}}}
\put(7051,-1936){\makebox(0,0)[lb]{\smash{\SetFigFont{8}{9.6}{\rmdefault}{\mddefault}{\updefault}Get $reach(a,e)$}}}
\put(1276,-2161){\makebox(0,0)[lb]{\smash{\SetFigFont{9}{10.8}{\rmdefault}{\mddefault}{\updefault}$N_9$:  $\leftarrow edge(a, X)$}}}
\put(4576,-2836){\makebox(0,0)[lb]{\smash{\SetFigFont{9}{10.8}{\rmdefault}{\mddefault}{\updefault}$N_{13}$:  $\Box$}}}
\put(1426,-1861){\makebox(0,0)[lb]{\smash{\SetFigFont{8}{9.6}{\rmdefault}{\mddefault}{\updefault}Get $reach(a,a)$}}}
\put(1576,-2836){\makebox(0,0)[lb]{\smash{\SetFigFont{9}{10.8}{\rmdefault}{\mddefault}{\updefault}$N_{10}$:  $\Box$}}}
\end{picture}

\caption{Answer iteration via a loop.}\label{fig4}
\end{figure}

The TP approach does backtracking in the same way as the OLDT
approach except for the following key differences:
(1) Because we do not distinguish between solution and lookup nodes/subgoals,
when no new answers in the table are available to
the subgoal $reach(a,Z)$ at $N_1$, we backtrack the subgoal by
resolving it against the next
clause $C_{r_3}$. This guarantees that TP-derivations are always linear.
(2) Since there is a loop between $N_0$ and $N_1$, 
before failing the subgoal $reach(a,X)$ at $N_0$ via backtracking we  
need to be sure that the subgoal has got its complete set
of answers. This is achieved
by performing {\em answer iteration} via the loop. 
That is, we regenerate the loop
to see if any new answers can be derived until we reach a
fixpoint. Figure \ref{fig3} shows the first part of TP-resolution,
where the following answers to $G_0$ are derived: $X=a$, 
$X=b$, $X=d$ and $X=e$. Figure \ref{fig4} shows the part of answer
iteration. Since no new answer is derived during the
iteration (i.e. no answer is added to any tables), 
we fail the subgoal $reach(a,X)$ at $N_0$.


\begin{remark}
{\em
From the above illustration, we see that 
in OLDT-resolution, solution nodes are those
at which the left-most subgoals are
generated earliest among all their
variant subgoals. In SLG-resolution, however, solution nodes are
roots of trees in a search forest, each labeled by
a special clause of the form $A \leftarrow A$ \cite{CSW95}. In
Tabulated SLS-resolution, any root of a tree in
a forest is itself labeled by
an instance, say $A \leftarrow B_1,...,B_n$ $(n \geq 0)$,
of a program clause and no nodes in the tree 
will produce child nodes using program clauses \cite{BD93}. 
However, for any atom $A$ we can
assume a virtual super-root labeled with $A \leftarrow A$,
which takes all the roots in the forest labeled by $A \leftarrow ...$ as its 
child nodes. In this sense, the search forest in Tabulated SLS-resolution
is the same as that in SLG-resolution for positive logic programs.
Therefore, we can consider all virtual super-roots as solution nodes.
}
\end{remark}

\section{TP-Resolution}

This section formally defines the TP approach to tabulated
resolution, mainly including the representation of tables,
the strategy for controlling tabulated derivations (TP-strategy), and 
the algorithm for making tabulated derivations based on
the control strategy (TP-trees).

\subsection{Tabled Predicates and Tables}

Predicates in a program $P$ are classified as {\em tabled
predicates} and {\em non-tabled predicates}. The classification is
made based on a {\em
dependency graph} \cite{ABW88}. Informally,
for any predicates $p$ and $q$, there is an edge
$p \rightarrow q$ in a dependency graph 
$G_P$ if there is a clause in $P$ of the form 
$p(.) \leftarrow ...,q(.),...$
Then a predicate $p$ is to be tabled if $G_P$
contains a cycle with a node $p$.

Any atom/subgoal with a tabled predicate is called a 
{\em tabled atom/subgoal}. During tabulated resolution,
we will create a table for each tabled subgoal, $A$. Apparently,
the table must contain $A$ (as an index) and have space to store
intermediate answers of $A$. Note that in our tabling approach,
any tabled subgoal can act both as a solution subgoal 
and as a lookup subgoal, so a table can be viewed as a 
blackboard on which a set of
variant subgoals will read and write
answers. In order to guarantee not losing answers
for any tabled subgoals (i.e. the table should contain all answers
that $A$ is supposed to have by applying its related clauses),
while avoiding redundant computations (i.e. after a clause has been used
by $A$, it should not be re-used by any other variant subgoal $A'$), 
a third component is needed in the table that keeps the status of
the clauses related to $A$. Therefore,
after a clause $C_i$ has been used by $A$, we change its status.
Then when evaluating a new subgoal $A'$ that
is a variant of $A$, $C_i$ will be ignored because all answers of 
$A$ derived via $C_i$ have already been stored in the table.
For any clause whose head is a tabled atom, its status can be 
{\em ``no longer available''} or {\em
``still available.''} We say that $C_i$ is 
{\em ``no longer available''} to $A$ if all answers of $A$ through
the application of $C_i$ have already been stored in the table of $A$. 
Otherwise, we say $C_i$ is {\em ``still available''} to $A$.
Finally, we need a flag variable $COMP$ in the table to 
indicate if all answers through the application
of all clauses related to $A$ have been completely stored in the table. 
This leads to the following.

\begin{definition}
\label{tb}
{\em
Let $P$ be a logic program and
$p(\vec{X})$ a tabled subgoal.
Let $P$ contain exactly $M$ clauses,
$C_{p_1},...,C_{p_M}$,
with a head $p(.)$.
A {\em table} for $p(\vec{X})$, denoted $TB(p(\vec{X}))$, is a four-tuple
$(p(\vec{X}),T,C,COMP)$, where 
\begin{enumerate}
\item
$T$ consists of tuples that are 
instances of $\vec{X}$, each $\vec{I}$ of which represents an
answer, $p(\vec{X})\vec{X}/\vec{I}$, to the subgoal. 

\item
$C$ is a vector of $M$ elements, with $C[i]=0$ (resp. $=1$)
representing that the status of $C_{p_i}$ w.r.t. $p(\vec{X})$
is {\em ``no longer available''} (resp. {\em ``still available''}).

\item
$COMP \in \{0,1\}$, with $COMP=1$ indicating that the answers of 
$p(\vec{X})$ have been completed. 
\end{enumerate}
}
\end{definition}

For convenience, we use $TB(p(\vec{X})) \rightarrow answer\_tuple[i]$
to refer to the $i$-th answer tuple in $T$,  
$TB(p(\vec{X})) \rightarrow clause\_status[i]$ to the status of $C_{p_i}$ w.r.t. 
$p(\vec{X})$, and $TB(p(\vec{X})) \rightarrow COMP$ to the flag $COMP$.

\begin{example}
\label{exp1}
{\em
Let $P$ be a logic program that contains exactly three clauses,
$C_{p_1},C_{p_2}$ and $C_{p_3}$, with a head $p(.)$. The table  

$\qquad TB(p(X, Y)):$ $(p(X,Y),$ $\{(a,b),(b,a),(b, c)\},(1,0,0),0)$

\noindent represents that there are three answers to 
$p(X,Y)$, namely 
$p(a,b)$, $p(b,a)$ and $p(b,c)$, and that $C_{p_2}$ 
and $C_{p_3}$ have already been
used by $p(X,Y)$ (or its variant subgoals) 
and $C_{p_1}$ is still available to $p(X,Y)$. 
Obviously, the answers of $p(X,Y)$ have not yet been completed.
The table

$\qquad TB(p(a, b)):$ $(p(a,b),$ $\{()\}, (0,1,1),1)$

\noindent shows that $p(a,b)$ has been proved true after applying
$C_{p_1}$. Note that since $p(a,b)$ contains no variables, its answer is
a 0-ary tuple. Finally, the table

$\qquad TB(p(a, X)):$ $(p(a,X),$ $\{\}, (0,0,0),1)$

\noindent represents that $p(a,X)$ has no answer at all.
}
\end{example}

Before introducing operations on tables, we define the
structure of nodes used in TP-resolution.

\begin{definition}
\label{reg}
{\em
Let $P$ be a logic program and
$G_i$ a goal $\leftarrow p(\vec{X}),A_2,...,A_m$.
By {\em ``register a node $N_i$ with} $G_i$'' we do the following:
(1) label $N_i$ with $G_i$, i.e. 
$N_i: \leftarrow p(\vec{X}),A_2,...,A_m$;
and (2) create the following structure for $N_i$: \\
\indent  $\bullet$ $answer\_ptr$, a pointer that points to an answer tuple in 
$TB(p(\vec{X}))$. \\
\indent $\bullet$ $clause\_ptr$, a pointer that points to a clause in $P$
with a head $p(.)$.\\
\indent $\bullet$ $clause\_SUSP$ (initially =0), a flag used for the 
update of clause status.\\
\indent $\bullet$ $node\_LOOP$ (initially =0), a flag showing 
if $N_i$ is a loop node.\\
\indent $\bullet$ $node\_ITER$ (initially =0), a flag showing 
if $N_i$ is an iteration node.\\
\indent $\bullet$ $node\_ANC$ (initially =$-1$), a flag showing if $N_i$ 
has any ancestor variant subgoals.

}
\end{definition}

For any field $F$ in the structure of $N_i$, we refer to it
by $N_i \rightarrow F$. The meaning of $N_i \rightarrow answer\_ptr$
and $N_i \rightarrow clause\_ptr$ is obvious. The remaining
fields will be defined by Definition \ref{loop} followed by
the procedure $nodetype\_update(.)$. 
We are now ready to define operations on tables.

\begin{definition}
\label{ops}
{\em
Let $P$ be a logic program with $M$ clauses with a head $p(.)$ and
$N_i$ a node labeled by a goal $\leftarrow p(\vec{X}),..., A_m.$
Let $NEW$ be a global flag variable used for answer iteration
(see Algorithm 2 for details).
We have the following basic operations on a table.
\begin{enumerate}

\item
$create(p(\vec{X}))$. Create a table $TB(p(\vec{X})):$
$(p(\vec{X}), T, C,COMP)$, with $T=\{\}$, $COMP=0$, and
$C[j]=1$ for all $1 \leq j \leq M$.

\item
$memo(p(\vec{X}), \vec{I})$, where $\vec{I}$ is 
an instance of $\vec{X}$. When $\vec{I}$ is not in $TB(p(\vec{X}))$,
add it to the end of the table, set $NEW=1$, and 
if $\vec{I}$ is a variant of $\vec{X}$, set 
$TB(p(\vec{X})) \rightarrow COMP=1$.

\item
$lookup(N_i,\vec{I_i})$. Fetch the next answer tuple in $TB(p(\vec{X}))$,
which is pointed by $N_i \rightarrow answer\_ptr$, into $\vec{I_i}$. 
If there is no next tuple, $\vec{I_i}=null$.

\item
$memo\_look(N_i,p(\vec{X}),\vec{I},\theta_i)$. It is a
compact operator, which combines $memo(.)$ and $lookup(.)$.
That is, it first performs $memo(p(\vec{X}), \vec{I})$ and
then gets the next answer tuple $\vec{F}$ from $TB(p(\vec{X}))$, 
which together with
$\vec{X}$ forms a substitution $\theta_i=\vec{X}/\vec{F}$.
If there is no next tuple, $\theta_i=null$.
\end{enumerate}
}
\end{definition}

First, the procedure $create(p(\vec{X}))$ is called only when the subgoal
$p(\vec{X})$ occurs the first time and no variant subgoals
occurred before. Therefore, up to the time when we call
$create(p(\vec{X}))$, no clauses with a head $p(.)$ in $P$
have been selected by any variant subgoals of $p(\vec{X})$,
so their status should be set to $1$. Second, whenever an answer 
$p(\vec{I})$ of $p(\vec{X})$ is derived, we call the
procedure $memo(p(\vec{X}),\vec{I})$. If the answer is new, it is 
appended to the end of the table. The flag $NEW$ is then set to $1$,
showing that a new answer has been derived. If the new tuple 
$\vec{I}$ is a variant of $\vec{X}$, which means that $p(\vec{X})$
is true for any instances of $\vec{X}$, the answers of $p(\vec{X})$ 
are completed so $TB(p(\vec{X})) \rightarrow COMP$ is set to 1. Finally, 
$lookup(N_i,\vec{I_i})$ is used to fetch an answer tuple from the table
for the subgoal $p(\vec{X})$ at $N_i$. 

$memo(.)$ and $lookup(.)$ can be used independently. They can also 
be used in pairs, i.e. $memo(.)$ immediately followed by $lookup(.)$.
In the latter case, it would be more convenient to use $memo\_look(.)$.

\subsection{TP-Strategy and TP-Trees}
In this subsection, we introduce the tabulated control
strategy and the way to make tabulated derivations based on
this strategy. We begin by discussing how to resolve subgoals
with program clauses and answers in tables.

Let $N_i$ be a node labeled by a goal
$G_i=\leftarrow A_1,...,A_m$
with $A_1=p(\vec{X})$ a tabled subgoal.
Consider evaluating $A_1$ using a program clause
$C_p= A \leftarrow B_1,...,B_n$ $(n \geq 0)$, 
where $A_1 \theta = A \theta$.\footnote{Here
and throughout, we assume that $C_p$ has been
standardized apart to share no variables with $G_i$.}
If we use SLD-resolution, we would obtain a new
node labeled with the goal 
$G_{i+1}=\leftarrow (B_1,...,B_n,A_2,...,A_m) \theta$,
where the mgu $\theta$ is {\em consumed} by all
$A_j$s $(j >1)$, although 
the proof of $A_1 \theta$ has not yet been completed ({\em
produced}). In order to avoid such kind of pre-consumption,
we propose the so called {\em PMF} (for Prove-Memorize-Fetch)
mode for resolving tabled subgoals with clauses. 
That is, we first prove $(B_1,...,B_n)\theta$. If it is true with
an mgu $\theta_1$, which means $A_1 \theta\theta_1$ is true,
we memorize the answer $A_1 \theta\theta_1$ in the table
$TB(A_1)$ if it is new. We then fetch an answer from 
$TB(A_1)$ to apply to the remaining subgoals of $G_i$.
Obviously modifying SLD-resolution by 
the PMF mode preserves the original answers to
$G_i$. Moreover, since only new answers are added to $TB(A_1)$,
all repeated answers of $A_1$ will be precluded to apply to
the remaining subgoals of $G_i$, so that redundant computations
are avoided.

The PMF mode can readily be realized by using the two table procedures,
$memo(.)$ and $lookup(.)$, or using the compact operator
$memo\_look(.)$. That is, 
after resolving the subgoal $A_1$ with the clause $C_p$,
$N_i$ gets a child node $N_{i+1}$ labeled with the goal 

$\quad G_{i+1}=\leftarrow (B_1,...,B_n)\theta,
memo\_look(N_i,p(\vec{X}),\vec{X}\theta,\theta_i),
A_2,...,A_m$.  

\noindent Note that the application of $\theta$ is blocked
by the subgoal $memo\_look(.)$ because the consumption
(fetch) must follow the production (prove and memorize).
We now explain how it works. 

Assume that after some resolution steps from $N_{i+1}$
we reach a node $N_k$ that is labeled by the goal
$G_k=\leftarrow memo\_look(N_i,p(\vec{X}),\vec{X}\theta\theta_1,\theta_i),
A_2,...,A_m$. This means that
$(B_1,...,B_n) \theta$ has been proved true
with the mgu $\theta_1$. That is, $A_1 \theta\theta_1$ is an
answer of $A_1$. By the left-most computation rule, 
$memo\_look(N_i,p(\vec{X}),\vec{X}\theta\theta_1,\theta_i)$ is executed, which adds
to the table $TB(A_1)$ the answer tuple $\vec{X}\theta\theta_1$ 
if it is new, gets from $TB(A_1)$ the next tuple $\vec{I}$,
and then sets $\theta_i=\vec{X}/\vec{I}$.
Since $A_1 \theta_i$ is an answer to the subgoal $A_1$ of $G_i$, the mgu
$\theta_i$ needs to be applied to the remaining $A_j$s of $G_i$.
We distinguish between two cases. 
\begin{enumerate}

\item[(1)] From $A_2$ to $A_m,$ $A_j=memo\_look(N_f,B,\_,\theta_f)$
is the first subgoal of the form $memo\_look(.)$. According to the PMF mode, 
there must be a node $N_f$, which occurred earlier than $N_i$, labeled with a goal
$G_f= \leftarrow B,A_{j+1},...,A_m$ such that $B$ is a tabled subgoal and
$A_j=memo\_look(N_f,B,\_,\theta_f)$ resulted from
resolving $B$ with a program clause. This means that
the proof of $B$ is now reduced to the proof of $(A_2,...,A_{j-1})\theta_i$.
Therefore, by the PMF mode $\theta_i$
should be applied to the subgoals $A_2$ until $A_j$. That is,
$N_k$ has a child node $N_{k+1}$ labeled with a goal
$G_{k+1}=\leftarrow (A_2,...,A_j)\theta_i,A_{j+1},...,A_m$.

\item[(2)] For no $j \geq 2$ $A_j$ is of the form $memo\_look(.)$. 
This means that no $A_j$ is a descendant of any tabled 
subgoal, so the mgu $\theta_i$ should be applied to all the $A_j$s. That is,
$G_{k+1}=\leftarrow (A_2,...,A_m)\theta_i$.
\end{enumerate}

Note that by Definition \ref{ops} the atom $p(\vec{X})$
in $memo(p(\vec{X}),\_)$ and $memo\_look(\_,p(\vec{X}),\_,\_)$ 
is merely used to index the table $TB(p(\vec{X}))$, so it cannot be 
instantiated during the resolution. That is, for any mgu $\theta$,
$memo(p(\vec{X}), \vec{I})\theta=memo(p(\vec{X}), \vec{I}\theta)$ and 
$memo\_look(N_i,p(\vec{X}),\vec{I},\theta_i)\theta=
memo\_look(N_i,p(\vec{X}),\vec{I}\theta,\theta_i)$

The above discussion shows how to resolve the tabled subgoal $A_1$
at $N_i$ against a program clause using the PMF mode. 
The same principle can be applied to
resolve $A_1$ with an answer tuple $\vec{I}$
in $TB(A_1)$ and to resolve $A_1$ with a program
clause when $A_1$ is a non-tabled subgoal.
Therefore, we have the following definition of resolvents for TP-resolution.

\begin{definition}
\label{sldtr}
{\em
Let $N_i$ be a node labeled by
a goal $G_i=\leftarrow A_1,...,A_m$ $(m \geq 1)$.

\begin{enumerate}
\item
If $A_1$ is $memo\_look(N_h,p(\vec{X}),\vec{I},\theta_h)$, then
the {\em resolvent} of $G_i$ and $\theta_h$ ($\theta_h \neq null$)
is the goal $G_{i+1}=\leftarrow (A_2,...,A_k)\theta_h,A_{k+1},...,A_m$,
where $A_k$ $(k>1)$ is 
the left-most subgoal of the form $memo\_look(.)$.

Otherwise, let $A_1=p(\vec{X})$ and
$C_p$ be a program clause
$A \leftarrow B_1,...,B_n$ with $A \theta = A_1 \theta$.

\item
If $A_1$ is a non-tabled subgoal, 
the {\em resolvent} of $G_i$ and $C_p$ is the goal
$G_{i+1}=\leftarrow (B_1,...,B_n,$
$A_2,...,A_k) \theta,$
$A_{k+1},...,A_m$, where  $A_k$ is 
the left-most subgoal of the form $memo\_look(.)$.

\item
If $A_1$ is a tabled subgoal, 
the {\em resolvent} of $G_i$ and $C_p$ is the goal
$G_{i+1}=\leftarrow (B_1,...,B_n)\theta,$
$memo\_look(N_i,p(\vec{X}),\vec{X}\theta,\theta_i),A_2,...,A_m$.

\item
If $A_1$ is a tabled subgoal, let $\vec{I}$ ($\vec{I} \neq null$)
be an answer tuple in $TB(A_1)$, then
the {\em resolvent} of $G_i$ and $\vec{I}$ is the goal
$G_{i+1}=\leftarrow (A_2,...,A_k)\vec{X}/\vec{I},A_{k+1},...,A_m$,
where  $A_k$ is 
the left-most subgoal of the form $memo\_look(.)$.
\end{enumerate}
}
\end{definition}

We now discuss tabulated control strategies. Recall that 
Prolog implements SLD-resolution by sequentially
searching an SLD-tree using the Prolog control strategy
({\em Prolog-strategy}, for short):
{\bf Depth-first} (for goal selection) + 
{\bf Left-most} (for subgoal selection)
+ {\bf Top-down} (for clause selection) + 
{\bf Last-first} (for backtracking). Let 
{\em ``register a node $N_i$ with $G_i$''}
be as defined by Definition
\ref{reg} except that the structure of $N_i$ only contains
the pointer $clause\_ptr$. Let
$return(\vec{Z})$ be a procedure that returns $\vec{Z}$
when $\vec{Z} \neq ()$ and {\em YES} otherwise.
Then the way that Prolog makes SLD-derivations based on
Prolog-strategy
can be formulated as follows.

\begin{definition}[Algorithm 1]
\label{prolog}
{\em
Let $P$ be a logic program and $G_0$ a top goal with
the list $\vec{Y}$ of variables.
The {\em Prolog-tree} $T_{G_0}$
of $P \cup \{G_0\}$ is constructed by recursively
performing the following steps until the answer $NO$ is returned.
\begin{enumerate}
\item
(Root node) Register the root $N_0$ with $G_0+return(\vec{Y})$ 
and goto 2.

\item
(Node expansion) Let $N_i$ be the latest registered
node labeled by $G_i= \leftarrow A_1,...,A_m$
$(i \geq 0, m>0)$.
Register $N_{i+1}$ as a child of $N_i$ with
$G_{i+1}$ if $G_{i+1}$ can be obtained as follows.

\begin{itemize}
\item
Case 1: $A_1$ is $return(.)$. Execute the procedure
$return(.)$, set $G_{i+1}=\Box$ (an empty clause),
and goto $3$ with $N=N_i$.

\item
Case 2: $A_1$ is an atom. Get a program clause
$A \leftarrow B_1,...,B_n$ (top-down via the pointer 
$N_i \rightarrow clause\_ptr$) such that $A_1 \theta = A \theta$.
If no such a clause exists, then goto $3$ with $N=N_i$; else
set $G_{i+1}=\leftarrow (B_1,...B_n,A_2,...,A_m) \theta$ and goto 2.

\end{itemize}

\item
(Backtracking)
If $N$ is the root, then return $NO$; else goto $2$ with
its parent node as the latest registered node.

\end{enumerate}
}
\end{definition}

Let $ST_{G_0}$ be the SLD-tree of $P \cup \{G_0\}$
via the left-most computation rule.\footnote{In \cite{TS86},
it is called an {\em OLD-tree}.} It is easy to prove that
when $P$ has the bounded-term-size property \cite{VG89} and
$ST_{G_0}$ contains no infinite loops,
Algorithm 1 is sound and 
complete in that $T_{G_0}=ST_{G_0}$. Moreover,
Algorithm 1 has the following distinct advantages:
(1) since SLD-resolution is linear, Algorithm 1
can be efficiently implemented using a simple
stack-based memory structure; (2) due to its
linearity and regular
sequentiality, some useful control mechanisms,
such as the well-known cut operator !, 
can be used to heuristically reduce search space.
Unfortunately, Algorithm 1 suffers from two serious
problems. One is that it is easy to get into infinite loops
even for very simple programs 
such as $P=\{p(X) \leftarrow p(X)\}$,
which makes it incomplete in many cases. The second problem 
is that it unnecessarily re-applies
the same set of clauses to variant subgoals
such as in the query $\leftarrow p(X),p(Y)$, which
leads to unacceptable performance.

As tabling has a distinct advantage of
resolving infinite loops and redundant derivations, 
one interesting
question then arises: Can we enhance Algorithm 1 with tabling, 
making it free from infinite loops and redundant computations
while preserving the above two advantages?
In the rest of this subsection, we give a constructive answer to
this question. We first discuss how to
enhance Prolog-strategy with tabling.

Observe that in a tabling system, we will
have both program clauses
and tables. For convenience, we refer to answer tuples in tables as 
{\em tabled facts}. Therefore, 
in addition to the existing policies in Prolog-strategy,
we need to have the following
two additional policies: (1) when both program clauses
and tabled facts are available, first use tabled facts
(i.e. {\bf Table-first} for program and table selection);
(2) when there are more than one tabled fact available,
first use the one that is earliest memorized.
Since we always add new answers to the end 
of tables (see Definition \ref{ops} for 
$memo(.)$), policy (2) amounts to saying {\bf Top-down}
selection for tabled facts.
This leads to the following control strategy for tabulated
derivations.

\begin{definition}
\label{tps}
{\em
By {\em TP-strategy} we mean:
Depth-first (for goal selection) + Left-most (for subgoal selection)
+ Table-first (for program and table selection)
+ Top-down (for the selection of tabled facts and program clauses) + 
Last-first (for backtracking).
}
\end{definition}

Our goal is to extend Algorithm 1 to make
linear tabulated derivations based on 
TP-strategy. To this end, we need to review a few concepts
concerning loop checking.

\begin{definition}[\cite{shen97} with slight modification]
\label{alist}
{\em
An {\it ancestor list} $AL_A$ of pairs $(N, B)$ is associated with 
each tabled subgoal $A$ at a node $N_i$ in a 
tree (see the TP-tree below), which
is defined recursively as follows.
\begin{enumerate}
\item
If $A$ is at the root, then $AL_A=\{\}$.

\item
If $A$ inherits a subgoal $A'$ (by copying or instantiation)
from its parent node, then $AL_A=AL_{A'}$.

\item
Let $A$ be in the resolvent of a subgoal $B$
at $N_f$ against a clause
$B' \leftarrow A_1, ..., A_n$ with 
$B \theta = B' \theta $ (i.e. $A=A_i \theta$ for some 
$1 \leq i \leq n$). 
If $B$ is a tabled subgoal, $AL_A=AL_B \cup \{(N_f,B)\}$;
otherwise $AL_A=\{\}$.
\end{enumerate}
}
\end{definition}

We see that for any tabled subgoals $A$ and $A'$, if 
$A$ is in the ancestor list of $A'$, i.e. $(\_,A) \in AL_{A'}$,
the proof of $A$ needs the proof of $A'$. Particularly,
if $(\_,A) \in AL_{A'}$ and $A'$ is a variant of $A$, the derivation
goes into a loop. This leads to the following.

\begin{definition}
\label{loop}
{\em
Let $G_i$ at $N_i$ and $G_k$ at $N_k$ be two goals 
in a derivation and
$A_i$ and $A_k$ be the left-most subgoals of $G_i$ and $G_k$,
respectively. We say $A_i$ (resp. $N_i$) is an 
{\em ancestor subgoal} of $A_k$ (resp. an {\em ancestor node} of $N_k$)
if $(N_i,A_i) \in AL_{A_k}$. If $A_i$ is both an ancestor subgoal
and a variant, i.e. an {\em ancestor variant subgoal}, 
of $A_k$, we say the derivation goes into a {\em loop},
denoted $L(N_i,N_k)$. Then,
$N_k$ and all its ancestor nodes involved in the loop are called
{\em loop nodes}. $N_i$ is also called the {\em top loop node} of
the loop. Finally, a loop node is called an {\em iteration node}
if by the time the node is about to fail through backtracking, 
it is the top loop node of all loops containing the node
that were generated before.
}
\end{definition}

\begin{figure}[h]
\centering
\setlength{\unitlength}{3947sp}%
\begingroup\makeatletter\ifx\SetFigFont\undefined%
\gdef\SetFigFont#1#2#3#4#5{%
  \reset@font\fontsize{#1}{#2pt}%
  \fontfamily{#3}\fontseries{#4}\fontshape{#5}%
  \selectfont}%
\fi\endgroup%
\begin{picture}(2942,3177)(5730,-4006)
\thicklines
\put(7126,-923){\circle{76}}
\put(7126,-1486){\circle{76}}
\put(7126,-2011){\circle{76}}
\put(7726,-2461){\circle{76}}
\put(6676,-2461){\circle{76}}
\put(7276,-2911){\circle{76}}
\put(7876,-3361){\circle{76}}
\put(8626,-2611){\circle{76}}
\put(7576,-3961){\circle{76}}
\put(5776,-2761){\circle{76}}
\put(7126,-1036){\vector( 0,-1){300}}
\put(7126,-1561){\vector( 0,-1){300}}
\put(7189,-2103){\vector( 3,-2){450}}
\put(7082,-2125){\vector(-4,-3){384}}
\put(6803,-2571){\vector( 3,-2){450}}
\put(7363,-2969){\vector( 3,-2){450}}
\multiput(6676,-2311)(28.84615,57.69231){14}{\makebox(6.6667,10.0000){\SetFigFont{10}{12}{\rmdefault}{\mddefault}{\updefault}.}}
\put(7801,-2461){\vector( 4,-1){741.177}}
\put(7876,-3436){\vector(-2,-3){300}}
\multiput(7501,-3886)(-15.95022,63.80089){14}{\makebox(6.6667,10.0000){\SetFigFont{10}{12}{\rmdefault}{\mddefault}{\updefault}.}}
\multiput(8551,-2536)(-64.12500,21.37500){21}{\makebox(6.6667,10.0000){\SetFigFont{10}{12}{\rmdefault}{\mddefault}{\updefault}.}}
\put(6601,-2536){\vector(-3,-1){742.500}}
\multiput(5851,-2686)(39.48387,52.64516){32}{\makebox(6.6667,10.0000){\SetFigFont{10}{12}{\rmdefault}{\mddefault}{\updefault}.}}
\put(7276,-961){\makebox(0,0)[lb]{\smash{\SetFigFont{10}{12.0}{\rmdefault}{\mddefault}{\updefault}$N_1$}}}
\put(7276,-1561){\makebox(0,0)[lb]{\smash{\SetFigFont{10}{12.0}{\rmdefault}{\mddefault}{\updefault}$N_2$}}}
\put(7276,-2011){\makebox(0,0)[lb]{\smash{\SetFigFont{10}{12.0}{\rmdefault}{\mddefault}{\updefault}$N_3$}}}
\put(7426,-2911){\makebox(0,0)[lb]{\smash{\SetFigFont{10}{12.0}{\rmdefault}{\mddefault}{\updefault}$N_4$}}}
\put(6076,-1786){\makebox(0,0)[lb]{\smash{\SetFigFont{10}{12.0}{\rmdefault}{\mddefault}{\updefault}$L_1$}}}
\put(7801,-2236){\makebox(0,0)[lb]{\smash{\SetFigFont{10}{12.0}{\rmdefault}{\mddefault}{\updefault}$L_3$}}}
\put(6976,-3511){\makebox(0,0)[lb]{\smash{\SetFigFont{10}{12.0}{\rmdefault}{\mddefault}{\updefault}$L_4$}}}
\put(6451,-2086){\makebox(0,0)[lb]{\smash{\SetFigFont{10}{12.0}{\rmdefault}{\mddefault}{\updefault}$L_2$}}}
\end{picture}

\caption{Loops, top loop nodes and iteration nodes.}\label{fig-loop}
\end{figure}
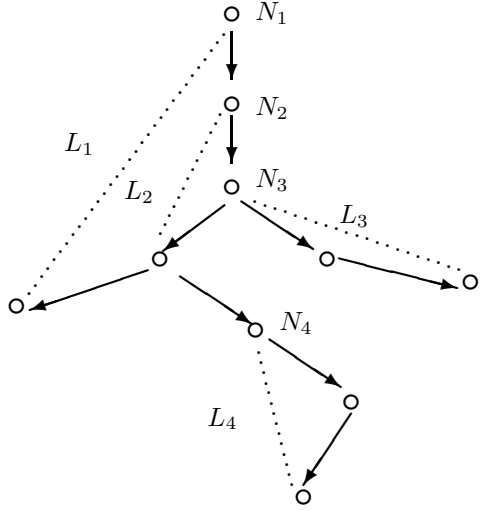

\begin{example}
\label{eg-loop}
{\em
Figure \ref{fig-loop} shows four loops, $L_1$, ..., $L_4$, 
with $N_1$, ..., $N_4$ their respective top loop nodes. We see that only 
$N_1$ and $N_4$ are iteration nodes.
}
\end{example}

Information about the types and ancestors of nodes is the basis on which we make
tabulated resolution. Such information
is kept in the structure of each node $N_i$ (see Definition \ref{reg}).
The flag $N_i \rightarrow node\_LOOP =1$ shows that $N_i$ is a loop node. The flag
$N_i \rightarrow node\_ITER =1$ shows that $N_i$ is an (candidate) iteration node. 
Let $A_1=p(\vec{X})$ be the left-most subgoal at $N_i$. The flag
$N_i \rightarrow node\_ANC=-1$ represents that it is unknown whether $A_1$ has 
any ancestor variant subgoal; $N_i \rightarrow node\_ANC=0$ shows 
that $A_1$ has no ancestor variant subgoal; and 
$N_i \rightarrow node\_ANC =j$ $(j>0)$ indicates that $A_1$
has ancestor variant subgoals and that $C_{p_j}$ is the clause
that is being used by its closest ancestor variant subgoal (i.e.,
let $A_h$ at $N_h$ be the closest ancestor variant subgoal of $A_1$, then
$N_i \rightarrow node\_ANC=j$ represents that $N_i$ is derived from $N_h$ 
via $C_{p_j}$).

Once a loop, say $L(N_1,N_m)$, of the form

$\qquad (N_1:\leftarrow A_1,...) \rightarrow_{{C_{p_j}},\theta_1} 
(N_2:\leftarrow A_2,...) \rightarrow ... 
\rightarrow (N_m:\leftarrow A_m,...)$

\noindent occurs, where all $N_i$s ($i<m$) are ancestor nodes of $N_m$ and  
$A_1=p(\vec{X})$ is the closest ancestor variant subgoal of $A_m$, 
we update the flags of all nodes, $N_1,...,N_m$, involved in the loop 
by calling the following procedure.

\begin{tabbing}
{\bf Procedure} $nodetype\_update(L(N_1,N_m))$ \\
$\quad$ \= (1) For all $i>1$ set $N_i \rightarrow node\_LOOP=1$
 and $N_i \rightarrow node\_ITER=0$. \\
\> (2) If $N_1 \rightarrow node\_LOOP=0$, set 
$N_1 \rightarrow node\_LOOP=1$ and
$N_1 \rightarrow node\_ITER=1$.\\
\> (3) Set $N_m \rightarrow node\_ANC=j$. \\ 
\> (4) For all $i<m$ set $N_i \rightarrow clause\_SUSP=1$. 
\end{tabbing} 

Point (1) is straightforward, where since $N_1$ is the top loop node of $L(N_1,N_m)$, 
all the remaining nodes in the loop cannot be an iteration node 
(see Definition \ref{loop}). 

If $N_1 \rightarrow node\_LOOP=0$,
meaning that $N_1$ is not involved in any loop that occurred before,
$N_1$ is considered as a {\em candidate} iteration node (point (2)).
A candidate iteration node becomes an iteration node if the node keeps its
candidacy by the time it is about to fail through backtracking 
(by that time it must be the top loop node of all 
previously generated loops containing it).

Since $A_1$ is the closest ancestor variant subgoal of $A_m$
and $C_{p_j}$ is the clause that is being used by $A_1$,
we set the flag $N_m \rightarrow node\_ANC=j$ (point (3)). 

As mentioned in Section 2, 
during TP-resolution when a loop $L(N_1,N_m)$ occurs, where the left-most 
subgoal $A_1=p(\vec{X})$ at $N_1$ is the closest ancestor 
variant subgoal of the left-most
subgoal $A_m$ at $N_m$, $A_m$ will skip the clause $C_{p_j}$ that
is being used by $A_1$. In order to ensure that such a skip will not
lead to loss of answers to $A_1$, we will do answer iteration before
failing $N_1$ via backtracking until we reach a fixpoint of answers.
Answer iteration is done by regenerating $L(N_1,N_m)$. 
This requires keeping the status of all clauses being used by the
loop nodes to {\em ``still available''} during backtracking.
Point (4) is used for such a purpose. After the flag
$N_i \rightarrow clause\_SUSP$ is set to 1, which indicates
that $N_i$ is currently involved in a loop, the status of the clause
being currently used by $N_i$ will not be set to {\em ``no longer
available''} when backtracking on $N_i$ (see Case B3 of Algorithm 2).     

\begin{remark}
{\em
We do answer iteration only at iteration nodes because they are
the top nodes of all loops involving them. If we did answer
iteration at a non-iteration loop node $N$, we would have to 
do it again at some top loop node $N_{top}$ of $N$, 
in order to reach a fixpoint at $N_{top}$ (see Figure \ref{fig-loop}). 
This would certainly lead to more redundant computations.  
}
\end{remark}

We are now in a position to define the TP-tree, which is
constructed based on the TP-strategy using the following algorithm.

\begin{definition}[Algorithm 2]
\label{sldt}
{\em
Let $P$ be a logic program and $G_0$ a top goal with the list $\vec{Y}$
of variables. The {\em TP-tree} $TP_{G_0}$ 
of $P \cup \{G_0\}$ is constructed by recursively
performing the following steps until the answer $NO$ is returned.
\begin{enumerate}
\item
(Root node) Register the root $N_0$ with $G_0+return(\vec{Y})$,
set $NEW=0$, and goto 2.

\item
(Node expansion) Let $N_i$ be the latest registered
node labeled by $G_i= \leftarrow A_1,...,A_m$
$(m>0)$. 
Register $N_{i+1}$ as a child of $N_i$ with
$G_{i+1}$ if $G_{i+1}$ can be obtained as follows.

\begin{itemize}
\item
Case 1: $A_1$ is $return(.)$. Execute the procedure
$return(.)$, set $G_{i+1}=\Box$ (an empty clause),
and goto $3$ with $N=N_i$.

\item
Case 2: $A_1$ is $memo\_look(N_h,p(\vec{X}),\vec{I},\theta_h)$. 
Execute the procedure.\footnote{See Definition \ref{ops}, where the flags
$NEW$ and $TB(p(\vec{X}))\rightarrow COMP$ will be updated.} 
If $\theta_h=null$ then goto $3$ with $N=N_i$; else set
$G_{i+1}$ to the resolvent of $G_i$ and $\theta_h$ and goto 2.

\item
Case 3: $A_1$ is a non-tabled subgoal. Get a clause $C$ 
whose head is unifiable with $A_1$.\footnote{Here and throughout,
clauses and answers in tables are selected top-down via the pointers
$N_i \rightarrow clause\_ptr$ and $N_i \rightarrow answer\_ptr$, respectively.} 
If no such a clause exists then goto $3$ with $N=N_i$; else
set $G_{i+1}$ to the resolvent of $G_i$ and $C$ and goto 2.

\item
Case 4: $A_1=p(\vec{X})$ is a tabled subgoal.  
Get an instance $\vec{I}$ of $\vec{X}$
from the table $TB(A_1)$.
If $\vec{I} \neq null$ then set $G_{i+1}$
to the resolvent of $G_i$ and $\vec{I}$ and goto 2.
Otherwise, if $TB(A_1) \rightarrow COMP=1$ then
goto 3 with $N=N_i$; else

\begin{itemize}
\item
Case 4.1: $N_i \rightarrow node\_ANC=-1$. 
If $A_1$ has no ancestor variant subgoal,
set $N_i \rightarrow node\_ANC=0$ and goto Case 4.2.
Otherwise, let $N_h$ be the closest ancestor node of $N_i$ 
such that $L(N_h,N_i)$ is a loop.
Call $nodetype\_update(L(N_h,N_i))$ and goto Case 4.3.

\item
Case 4.2: $N_i \rightarrow node\_ANC=0$. Get a clause $C_{p_j}$
whose head is unifiable with $A_1$ such that
$TB(A_1) \rightarrow clause\_status[j]=1$. If such a clause exists,
set $G_{i+1}$ to the resolvent of $G_i$ and $C_{p_j}$ and goto 2.
Otherwise, if $N_i \rightarrow node\_ITER=0$ then 
goto 3 with $N=N_i$; else
\begin{itemize}
\item
Case 4.2.1: $NEW=0$. Set $TB(A_1) \rightarrow COMP=1$ and 
goto $3$ with $N=N_i$. 
\item
Case 4.2.2: $NEW=1$. Set $NEW=0$, reset
$N_i \rightarrow clause\_ptr$ to pointing to the first clause 
$C_{p_j}$ whose status
is {\em ``still available''}, and goto Case 4.2. 
\end{itemize}

\item
Case 4.3: $N_i \rightarrow node\_ANC=j$ $(j>0)$. 
Get a clause $C_{p_k}$ ($k>j$)
whose head is unifiable with $A_1$ such that
$TB(A_1) \rightarrow clause\_status[k]=1$. If such a clause exists
then set $G_{i+1}$ to the resolvent
of $G_i$ and $C_{p_k}$ and goto 2; else goto $3$ with $N=N_i$.  

\end{itemize}
\end{itemize}

\item
(Backtracking) 
If $N$ is the root, return $NO$. Otherwise let $N_f$ be 
the parent node of $N$ with the left-most subgoal $A_f$.
\begin{itemize}
\item
Case B1: $A_f$ is $memo\_look(.)$. Goto $3$ with $N=N_f$.

\item
Case B2: $A_f$ is a non-tabled subgoal. 
Goto $2$ with $N_f$ as the latest registered node.

\item
Case B3: $A_f=q(\vec{Z})$ is a tabled subgoal. Let $N$ be generated from $N_f$
by resolving $A_f$ with a clause $C_{q_j}$. 
If $N_f \rightarrow node\_SUSP=0$ then set 
$TB(A_f) \rightarrow clause\_status[j]=0$; else set
$N_f \rightarrow node\_SUSP=0$.
Goto $2$ with $N_f$ as the latest registered node.
\end{itemize}
\end{enumerate}
}
\end{definition}

Obviously, Algorithm 2 reduces to Algorithm 1
when $P$ contains no tabled predicates.
We now explain Algorithm 2 briefly.
First we set up the root $N_0$ via registration (see
Definition \ref{reg}). The global variable $NEW$
is initialized to $0$, meaning that up to now
no new answer has been derived for any subgoal. 
Then by the Depth-first policy we select the
latest registered node, say $N_i$ labeled with the goal
$G_i$, for expansion (point 2). If the left-most subgoal $A_1$
of $G_i$ is $return(\vec{I})$ (Case 1), which means the top goal $G_0$
has been proved true with the answer substitution
$\vec{Y}/\vec{I}$, we reach a success leaf $N_{i+1}$ labeled
with $\Box$. We then do
backtracking (point 3) to derive alternative answers to $G_0$.

If $A_1$ is $memo\_look(N_h,p(\vec{X}),\vec{I},\theta_h)$ (Case 2), 
which means that the left-most subgoal
$p(\vec{X})$ at node $N_h$ is proved true with the answer 
substitution $\vec{X}/\vec{I}$,
we memorize $\vec{I}$ in the table $TB(p(\vec{X}))$ and set $NEW=1$ if the
answer is new. Meanwhile, if the answer $p(\vec{I})$ is
a variant of the subgoal $p(\vec{X})$, we set the flag
$TB(p(\vec{X})) \rightarrow COMP=1$, indicating that
the answers of $p(\vec{X})$ have been completed. After memorization,
we fetch the next answer 
from the table and then prove
the resolvent $G_{i+1}$ of $G_i$ and the new answer.

Case 3 is straightforward, so we move to Case 4.
By the Table-first policy, we first look up answers for
$A_1$ from the table
$TB(p(\vec{X}))$. When available, we fetch the next unused answer for
$A_1$ and create
the resolvent $G_{i+1}$. Otherwise,
we check the flag $TB(p(\vec{X})) \rightarrow COMP$ 
to see if the answers of $p(\vec{X})$ 
have been completed. If yes, which means that the subgoal $A_1$ at
$N_i$ has used all its answers, we backtrack to its parent node.
Otherwise, we continue to derive new answers by resolving
$A_1$ with the remaining clauses. Based on whether $A_1$ has any ancestor
variant subgoal, we distinguish three cases (Cases 4.1, 4.2 and 4.3).

At the time that $N_i$ is registered (see 
Definition \ref{reg}), we do not know if $A_1$ at
$N_i$ has any ancestor variant subgoal (i.e. $N_i \rightarrow node\_ANC=-1$
initially). So we check it via the ancestor list 
$AL_{A_1}$ (see Definition \ref{alist}) and update the flag 
$N_i \rightarrow node\_ANC$ accordingly. If $N_h$ is the closest
ancestor node of $N_i$ such that $L(N_h,N_i)$ is a loop, the other flags of
$N_i$, namely $node\_LOOP$, $node\_ITER$ and $node\_SUSP$, will also
be updated by the procedure $nodetype\_update(.)$ (see Case 4.1).

For Case 4.2, $A_1$ has no ancestor variant subgoal, 
which implies that the derivation does not
get into a loop at $N_i$. So we seek the next clause
whose head is unifiable with $A_1$ and whose status
is {\em ``still available,''} and use it to build  
the resolvent $G_{i+1}$. Now consider the case that
no such a clause exists, which means that the subgoal $A_1$ at
$N_i$ has used all its answers and clauses available.
In Prolog, we would fail the subgoal immediately and 
backtrack to its parent node. In TP-resolution, however,
we cannot do this unless $N_i$ is a non-iteration node.
Suppose $N_i$ is an iteration node (i.e. $N_i \rightarrow node\_ITER=1$).
Before failing $A_1$ via backtracking, we do answer iteration to
complete its answers. The process is quite simple. We
start an iteration simply by initializing $NEW$ to 0 and resetting the
pointer $N_i \rightarrow clause\_ptr$ to pointing to the
first clause $C_{p_j}$ whose status
remains to {\em ``still available''} (Case 4.2.2). 
Since the status of all clauses involved
in loops are kept to {\em ``still available''} during backtracking,
all the loops can be regenerated by the iteration.
By the end of an iteration, i.e. when we come back to $N_i$ again
and try to fail $A_1$ via backtracking, we check the flag $NEW$
to see if the termination condition is satisfied. If
$NEW=0$, meaning that a fixpoint has been reached
so that the answers of $A_1$ have been completed, we
stop answer iteration by failing $A_1$ via backtracking (see
Cases 4.2.1). Otherwise, we start a new iteration to
seek more answers (Case 4.2.2). 

For Case 4.3, $A_1$ has an ancestor variant subgoal, 
so the derivation has gone into a loop, 
say $L(N_h,N_i)$. In order to break 
the loop, we skip the clause $C_{p_j}$ that is being used by the
top loop node $N_h$. The skip of clauses may lead to loss of answers, 
which is the only reason why answer iteration is required.
({\bf Remark:} Algorithm 2 uses loop checking to cut loops 
and adopts answer iteration to iteratively regenerate loops that
are pruned by loop checking. Such a complementary use of loop checking and 
answer iteration is an effective way of cutting infinite loops
while guaranteeing the completeness of answers.)

Backtracking (point 3) is done as usual except that 
the status of the clauses that have been used should be set to
{\em ``no longer available''} (Case B3). Let $C_{q_j}$ be the 
clause that is being used by $N_f$. If no loop occurred that
went through $N_f$ via $C_{q_j}$, the flag $N_f \rightarrow node\_SUSP$
must remain to 0. In this case, we set the status of $C_{q_j}$
in $TB(A_f)$ to {\em ``no longer available''} 
because all answers of $A_f$ by the application
of $C_{q_j}$ have been exhausted. Otherwise, when a loop occurred before
that went through $N_f$ via $C_{q_j}$, $N_f \rightarrow node\_SUSP$ must
be 1 (see the procedure $nodetype\_update(.))$. So we keep the status
of $C_{q_j}$ to {\em ``still available''} while setting 
$N_f \rightarrow node\_SUSP$ to 0 again. 

Based on TP-trees, we have the following standard
definitions.
\begin{definition}
\label{tpr}
{\em
Let $TP_{G_0}$ be a TP-tree
of $P \cup \{G_0\}$.
All leaves of $TP_{G_0}$ labeled by
$\Box$ are {\em success} leaves and all other
leaves are {\em failure} leaves.
A {\em TP-derivation},
denoted by
$G_0 \Rightarrow_{C_1, \theta_1} G_1\Rightarrow ... $
$\Rightarrow_{C_i,\theta_i}G_i \Rightarrow ...$
$\Rightarrow_{C_n, \theta_n} G_n$, is a partial
branch in $TP_{G_0}$
starting at the root, where each $G_i$ is a goal labeling
a node $N_i$ and for each $0 \leq i<n$,
$G_{i+1}$ is the resolvent of $G_i$ and $C_{i+1}$ with
the mgu $\theta_{i+1}$, where $C_{i+1}$ may be a program clause
or a tabled fact or blank (when the left-most subgoal
of $G_i$ is a procedure).
A TP-derivation is {\em successful} if it ends with a
success leaf and {\em failed}, otherwise.
The process of constructing TP-derivations
is called {\em TP-resolution}.
}
\end{definition}

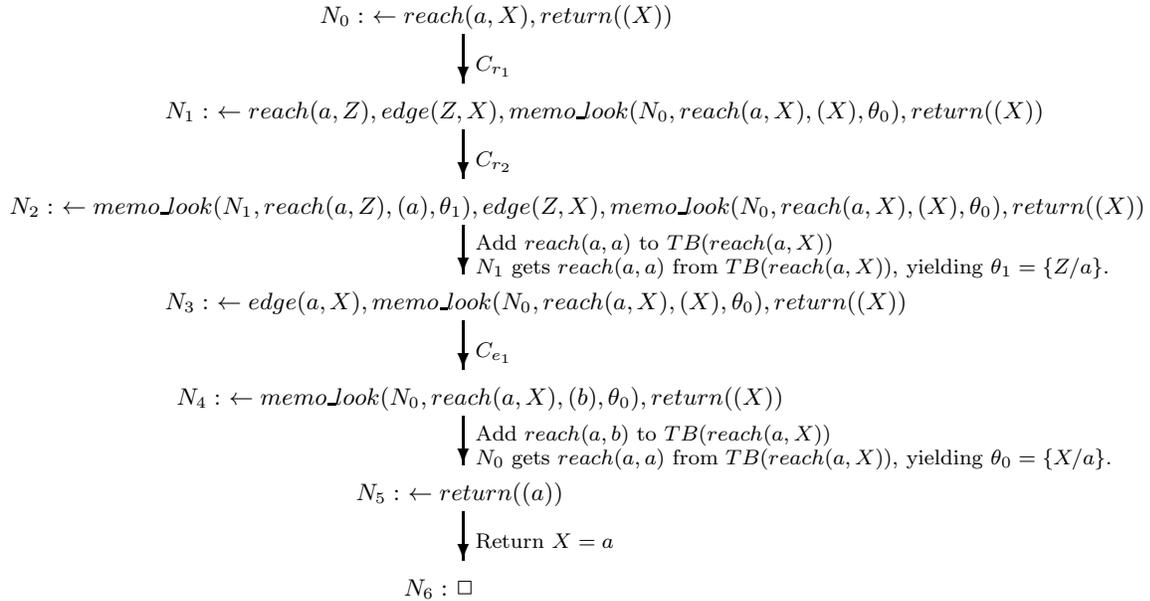
\begin{figure}[h]
\setlength{\unitlength}{3947sp}%
\begingroup\makeatletter\ifx\SetFigFont\undefined%
\gdef\SetFigFont#1#2#3#4#5{%
  \reset@font\fontsize{#1}{#2pt}%
  \fontfamily{#3}\fontseries{#4}\fontshape{#5}%
  \selectfont}%
\fi\endgroup%
\begin{picture}(2925,3705)(1276,-3832)
\thicklines
\put(4126,-286){\vector( 0,-1){300}}
\put(4126,-886){\vector( 0,-1){300}}
\put(4126,-1486){\vector( 0,-1){300}}
\put(4126,-2086){\vector( 0,-1){300}}
\put(4126,-2686){\vector( 0,-1){300}}
\put(4126,-3286){\vector( 0,-1){300}}
\put(2326,-2611){\makebox(0,0)[lb]{\smash{\SetFigFont{9}{10.8}{\rmdefault}{\mddefault}{\updefault}$N_4:$  $\leftarrow memo\_look(N_0,reach(a,X),(b),\theta_0),return((X))$}}}
\put(3451,-3211){\makebox(0,0)[lb]{\smash{\SetFigFont{9}{10.8}{\rmdefault}{\mddefault}{\updefault}$N_5:$  $\leftarrow return((a))$}}}
\put(4201,-1111){\makebox(0,0)[lb]{\smash{\SetFigFont{8}{9.6}{\rmdefault}{\mddefault}{\updefault}$C_{r_2}$}}}
\put(4201,-1636){\makebox(0,0)[lb]{\smash{\SetFigFont{8}{9.6}{\rmdefault}{\mddefault}{\updefault}Add $reach(a,a)$ to $TB(reach(a,X))$}}}
\put(4201,-1786){\makebox(0,0)[lb]{\smash{\SetFigFont{8}{9.6}{\rmdefault}{\mddefault}{\updefault}$N_1$ gets $reach(a,a)$ from $TB(reach(a,X))$, yielding $\theta_1=\{Z/a\}$.}}}
\put(4201,-2311){\makebox(0,0)[lb]{\smash{\SetFigFont{8}{9.6}{\rmdefault}{\mddefault}{\updefault}$C_{e_1}$}}}
\put(4201,-3511){\makebox(0,0)[lb]{\smash{\SetFigFont{8}{9.6}{\rmdefault}{\mddefault}{\updefault}Return $X=a$}}}
\put(3751,-3811){\makebox(0,0)[lb]{\smash{\SetFigFont{9}{10.8}{\rmdefault}{\mddefault}{\updefault}$N_6:$   $\Box$}}}
\put(4201,-2986){\makebox(0,0)[lb]{\smash{\SetFigFont{8}{9.6}{\rmdefault}{\mddefault}{\updefault}$N_0$ gets $reach(a,a)$ from $TB(reach(a,X))$, yielding $\theta_0=\{X/a\}$.}}}
\put(4201,-2836){\makebox(0,0)[lb]{\smash{\SetFigFont{8}{9.6}{\rmdefault}{\mddefault}{\updefault}Add $reach(a,b)$ to $TB(reach(a,X))$}}}
\put(3226,-211){\makebox(0,0)[lb]{\smash{\SetFigFont{9}{10.8}{\rmdefault}{\mddefault}{\updefault}$N_0:$  $\leftarrow reach(a,X),return((X))$ }}}
\put(2251,-811){\makebox(0,0)[lb]{\smash{\SetFigFont{9}{10.8}{\rmdefault}{\mddefault}{\updefault}$N_1:$  $\leftarrow reach(a,Z),edge(Z,X),memo\_look(N_0,reach(a,X),(X),\theta_0),return((X))$}}}
\put(4201,-511){\makebox(0,0)[lb]{\smash{\SetFigFont{8}{9.6}{\rmdefault}{\mddefault}{\updefault}$C_{r_1}$}}}
\put(1276,-1411){\makebox(0,0)[lb]{\smash{\SetFigFont{9}{10.8}{\rmdefault}{\mddefault}{\updefault}$N_2:$  $\leftarrow memo\_look(N_1,reach(a,Z),(a),\theta_1),edge(Z,X),memo\_look(N_0,reach(a,X),(X),\theta_0),return((X))$}}}
\put(2251,-2011){\makebox(0,0)[lb]{\smash{\SetFigFont{9}{10.8}{\rmdefault}{\mddefault}{\updefault}$N_3:$  $\leftarrow edge(a,X),memo\_look(N_0,reach(a,X),(X),\theta_0),return((X))$}}}
\end{picture}

\caption{The first successful TP-derivation with an answer $X=a$.}\label{fig-reach1}
\end{figure}

\begin{example}
\label{exp2}
{\em
Consider the example program $P_1$ again (see Section 2). 
Based on the dependency graph of $P_1$, we choose $reach$ 
as a tabled predicate and $edge$ as a non-tabled one. 
Now consider applying Algorithm 2 to the top goal 
$G_0= \leftarrow reach(a, X)$. 

We first set up the root $N_0$ labeled with
$\leftarrow reach(a, X),return((X))$ and set $NEW=0$ (point 1). 
Then we expand $N_0$ using the clause $C_{r_1}$ (point 2, Cases 4, 
4.1 and 4.2), which creates a table

$\quad TB(reach(a,X)): (reach(a,X), \{\}, (1,1,1),0)$

\noindent and a child node $N_1$ (see Figure \ref{fig-reach1}).
Obviously there is a loop $L(N_0,N_1)$, so we call the procedure
$notetype\_update(L(N_0,N_1))$, which marks $N_0$ as a candidate
iteration node, sets $N_0 \rightarrow clause\_SUSP=1$ and
$N_1 \rightarrow node\_ANC=1$. Then by Case 4.3 the clause $C_{r_2}$
(instead of $C_{r_1}$) is applied to $reach(a, Z)$ at $N_1$, 
which gives a node $N_2$. Next, by Case 2
the answer $reach(a,a)$ is memorized in the table (so $NEW=1$), yielding

$\quad TB(reach(a, X)): (reach(a, X), \{(a)\}, (1,1,1),0)$

\noindent and the node $N_3$ is derived using the first tabled
fact. By successively performing Cases 3, 2 and 1,
we reach a success leaf $N_6$ with the first answer $X=a$ to the top
goal. After these steps, the table looks like

$\quad TB(reach(a, X)): (reach(a, X), \{(a),(b)\}, (1,1,1),0)$.

Now we do backtracking. By Cases B1 and B2 we go back
until $N_3$. Since $C_{e_2}$ is not unifiable with the subgoal
$edge(a,X)$, we go back to $N_2$ and then to
$N_1$. From $N_1$ we consecutively
derive a failure leaf $N_7$ (Figure \ref{fig-reach2}), 
a success leaf $N_{12}$
(Figure \ref{fig-reach3}) and another failure leaf $N_{13}$ 
(Figure \ref{fig-reach4}).
After these steps, the table becomes

$\quad TB(reach(a,X)): (reach(a,X), \{(a),(b),(d),(e)\}, (1,0,0),0)$.

Now $reach(a,Z)$ at 
$N_1$ has used all answers in the table and has no more clause available. 
So we return to the root $N_0$. Note that since the flag
$N_0 \rightarrow clause\_SUSP=1$, which shows the clause $C_{r_1}$
that is being used by $N_0$ is involved in a loop, the status of
$C_{r_1}$ in $TB(reach(a,X))$ remains to {\em ``still available''}
when backtracking from $N_1$ to $N_0$ (see Case B3). 

From Figures \ref{fig-reach1}$-$\ref{fig-reach4}, we
see that $N_0$ has used only the first two answers in $TB(reach(a,X))$,
namely $reach(a,a)$ and $reach(a,b)$. So it continues to use the other   
two. By repeating Case 4, 
Case 1 and point 3 twice, we get another two successful derivations
as depicted in Figures \ref{fig-reach5} and \ref{fig-reach6}. 

Now $reach(a,X)$ at $N_0$ has used all tabled facts in $TB(reach(a,X))$
and has no more clause available (note that $C_{r_2}$ and $C_{r_3}$ 
have already been used by $N_1$). Before failing it via backtracking,
we check if $N_0$ is an iteration node (i.e. we see if 
$N_0 \rightarrow node\_ITER$ remains to the value 1). Since $N_0$ is an iteration
node and the flag $NEW=1$, by Case 4.2.2 we do answer iteration. 
It is easy to check that no new answer will be derived (see
Figure \ref{fig4}), so by
the end of the first iteration $NEW$ remains to the value $0$. Thus 
by Case 4.2.1, the flag $COMP$ of $TB(reach(a,X))$ is changed to 1,
showing that the answers of $reach(a,X)$ have been completed.

Finally, by point 3 the answer $NO$ is returned, which
terminates the algorithm. Therefore by putting together 
Figures \ref{fig-reach1}$-$\ref{fig-reach6}
and the figures for answer iteration (which are omitted here)
we obtain the TP-tree $TP_{G_0}$ of $P_1 \cup \{G_0\}$.
}
\end{example}

\begin{figure}[h]
\setlength{\unitlength}{3947sp}%
\begingroup\makeatletter\ifx\SetFigFont\undefined%
\gdef\SetFigFont#1#2#3#4#5{%
  \reset@font\fontsize{#1}{#2pt}%
  \fontfamily{#3}\fontseries{#4}\fontshape{#5}%
  \selectfont}%
\fi\endgroup%
\begin{picture}(2400,969)(1501,-1510)
\thicklines
\put(3826,-961){\vector( 0,-1){300}}
\put(3901,-1261){\makebox(0,0)[lb]{\smash{\SetFigFont{8}{9.6}{\rmdefault}{\mddefault}{\updefault}$N_1$ gets $reach(a,b)$ from $TB(reach(a,X))$ with mgu $\{Z/b\}$}}}
\put(3901,-1111){\makebox(0,0)[lb]{\smash{\SetFigFont{8}{9.6}{\rmdefault}{\mddefault}{\updefault}The status of $C_{r_2}$ becomes {\em "no longer available"}}}}
\put(1576,-1486){\makebox(0,0)[lb]{\smash{\SetFigFont{9}{10.8}{\rmdefault}{\mddefault}{\updefault}$\qquad$ $N_7:$ $\leftarrow edge(b,X),memo\_look(N_0,reach(a,X),(X),\theta_0),return((X))$}}}
\put(1501,-886){\makebox(0,0)[lb]{\smash{\SetFigFont{9}{10.8}{\rmdefault}{\mddefault}{\updefault}$\qquad$ $N_1:$ $\leftarrow reach(a,Z),edge(Z,X),memo\_look(N_0,reach(a,X),(X),\theta_0),return((X))$}}}
\put(2176,-661){\makebox(0,0)[lb]{\smash{\SetFigFont{10}{12.0}{\rmdefault}{\mddefault}{\updefault}$\qquad$}}}
\end{picture}

\caption{A failed TP-derivation.
$\qquad\qquad\qquad\qquad\qquad\qquad\qquad\qquad$ }\label{fig-reach2}

\end{figure}

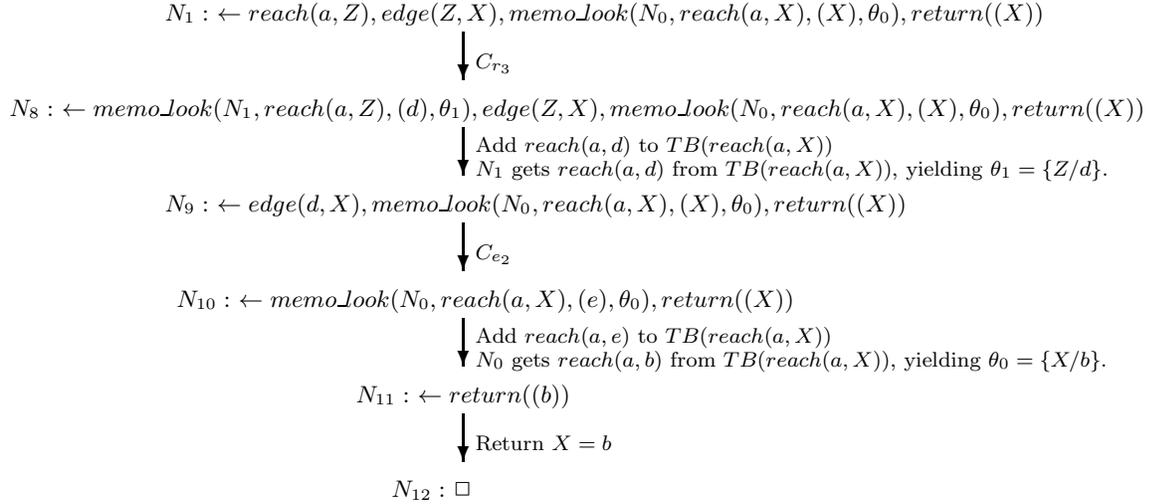
\begin{figure}[p]
\setlength{\unitlength}{3947sp}%
\begingroup\makeatletter\ifx\SetFigFont\undefined%
\gdef\SetFigFont#1#2#3#4#5{%
  \reset@font\fontsize{#1}{#2pt}%
  \fontfamily{#3}\fontseries{#4}\fontshape{#5}%
  \selectfont}%
\fi\endgroup%
\begin{picture}(2925,3477)(1276,-3832)
\thicklines
\put(4126,-886){\vector( 0,-1){300}}
\put(4126,-1486){\vector( 0,-1){300}}
\put(4126,-2086){\vector( 0,-1){300}}
\put(4126,-2686){\vector( 0,-1){300}}
\put(4126,-3286){\vector( 0,-1){300}}
\put(2326,-2611){\makebox(0,0)[lb]{\smash{\SetFigFont{9}{10.8}{\rmdefault}{\mddefault}{\updefault}$N_{10}:$  $\leftarrow memo\_look(N_0,reach(a,X),(e),\theta_0),return((X))$}}}
\put(3451,-3211){\makebox(0,0)[lb]{\smash{\SetFigFont{9}{10.8}{\rmdefault}{\mddefault}{\updefault}$N_{11}:$  $\leftarrow return((b))$}}}
\put(4201,-1111){\makebox(0,0)[lb]{\smash{\SetFigFont{8}{9.6}{\rmdefault}{\mddefault}{\updefault}$C_{r_3}$}}}
\put(4201,-1636){\makebox(0,0)[lb]{\smash{\SetFigFont{8}{9.6}{\rmdefault}{\mddefault}{\updefault}Add $reach(a,d)$ to $TB(reach(a,X))$}}}
\put(4201,-1786){\makebox(0,0)[lb]{\smash{\SetFigFont{8}{9.6}{\rmdefault}{\mddefault}{\updefault}$N_1$ gets $reach(a,d)$ from $TB(reach(a,X))$, yielding $\theta_1=\{Z/d\}$.}}}
\put(4201,-2311){\makebox(0,0)[lb]{\smash{\SetFigFont{8}{9.6}{\rmdefault}{\mddefault}{\updefault}$C_{e_2}$}}}
\put(4201,-3511){\makebox(0,0)[lb]{\smash{\SetFigFont{8}{9.6}{\rmdefault}{\mddefault}{\updefault}Return $X=b$}}}
\put(4201,-2986){\makebox(0,0)[lb]{\smash{\SetFigFont{8}{9.6}{\rmdefault}{\mddefault}{\updefault}$N_0$ gets $reach(a,b)$ from $TB(reach(a,X))$, yielding $\theta_0=\{X/b\}$.}}}
\put(4201,-2836){\makebox(0,0)[lb]{\smash{\SetFigFont{8}{9.6}{\rmdefault}{\mddefault}{\updefault}Add $reach(a,e)$ to $TB(reach(a,X))$}}}
\put(1276,-1411){\makebox(0,0)[lb]{\smash{\SetFigFont{9}{10.8}{\rmdefault}{\mddefault}{\updefault}$N_8:$  $\leftarrow memo\_look(N_1,reach(a,Z),(d),\theta_1),edge(Z,X),memo\_look(N_0,reach(a,X),(X),\theta_0),return((X))$}}}
\put(2251,-2011){\makebox(0,0)[lb]{\smash{\SetFigFont{9}{10.8}{\rmdefault}{\mddefault}{\updefault}$N_9:$  $\leftarrow edge(d,X),memo\_look(N_0,reach(a,X),(X),\theta_0),return((X))$}}}
\put(2701,-511){\makebox(0,0)[lb]{\smash{\SetFigFont{12}{14.4}{\rmdefault}{\mddefault}{\updefault}$\qquad$}}}
\put(2251,-811){\makebox(0,0)[lb]{\smash{\SetFigFont{9}{10.8}{\rmdefault}{\mddefault}{\updefault}$N_1:$  $\leftarrow reach(a,Z),edge(Z,X),memo\_look(N_0,reach(a,X),(X),\theta_0),return((X))$}}}
\put(3676,-3811){\makebox(0,0)[lb]{\smash{\SetFigFont{9}{10.8}{\rmdefault}{\mddefault}{\updefault}$N_{12}:$   $\Box$}}}
\end{picture}

\caption{The second successful TP-derivation with the second answer 
$X=b$.}\label{fig-reach3}
\end{figure}
 
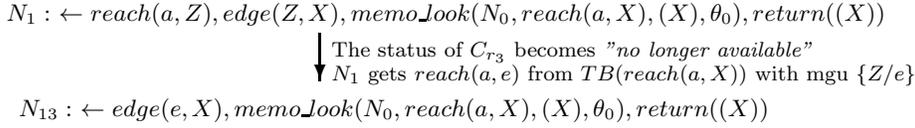
\begin{figure}[p]

\setlength{\unitlength}{3947sp}%
\begingroup\makeatletter\ifx\SetFigFont\undefined%
\gdef\SetFigFont#1#2#3#4#5{%
  \reset@font\fontsize{#1}{#2pt}%
  \fontfamily{#3}\fontseries{#4}\fontshape{#5}%
  \selectfont}%
\fi\endgroup%
\begin{picture}(2400,969)(1501,-1510)
\thicklines
\put(3826,-961){\vector( 0,-1){300}}
\put(3901,-1261){\makebox(0,0)[lb]{\smash{\SetFigFont{8}{9.6}{\rmdefault}{\mddefault}{\updefault}$N_1$ gets $reach(a,e)$ from $TB(reach(a,X))$ with mgu $\{Z/e\}$}}}
\put(3901,-1111){\makebox(0,0)[lb]{\smash{\SetFigFont{8}{9.6}{\rmdefault}{\mddefault}{\updefault}The status of $C_{r_3}$ becomes {\em "no longer available"}}}}
\put(1576,-1486){\makebox(0,0)[lb]{\smash{\SetFigFont{9}{10.8}{\rmdefault}{\mddefault}{\updefault}$\qquad$ $N_{13}:$ $\leftarrow edge(e,X),memo\_look(N_0,reach(a,X),(X),\theta_0),return((X))$}}}
\put(1501,-886){\makebox(0,0)[lb]{\smash{\SetFigFont{9}{10.8}{\rmdefault}{\mddefault}{\updefault}$\qquad$ $N_1:$ $\leftarrow reach(a,Z),edge(Z,X),memo\_look(N_0,reach(a,X),(X),\theta_0),return((X))$}}}
\put(2176,-661){\makebox(0,0)[lb]{\smash{\SetFigFont{10}{12.0}{\rmdefault}{\mddefault}{\updefault}$\qquad$}}}
\end{picture}

\caption{Another failed TP-derivation.
$\qquad\qquad\qquad\qquad\qquad\qquad\qquad\qquad$}\label{fig-reach4}
\end{figure}

\begin{figure}[p]
\setlength{\unitlength}{3947sp}%
\begingroup\makeatletter\ifx\SetFigFont\undefined%
\gdef\SetFigFont#1#2#3#4#5{%
  \reset@font\fontsize{#1}{#2pt}%
  \fontfamily{#3}\fontseries{#4}\fontshape{#5}%
  \selectfont}%
\fi\endgroup%
\begin{picture}(2400,1602)(1801,-1432)
\thicklines
\put(4051,-286){\vector( 0,-1){300}}
\put(4051,-886){\vector( 0,-1){300}}
\put(1951, 14){\makebox(0,0)[lb]{\smash{\SetFigFont{12}{14.4}{\rmdefault}{\mddefault}{\updefault}$\qquad$}}}
\put(1801,-211){\makebox(0,0)[lb]{\smash{\SetFigFont{9}{10.8}{\rmdefault}{\mddefault}{\updefault}$\qquad\qquad\qquad\qquad$ $N_0:$ $\leftarrow reach(a,X),return((X))$ }}}
\put(4201,-436){\makebox(0,0)[lb]{\smash{\SetFigFont{8}{9.6}{\rmdefault}{\mddefault}{\updefault}$N_0$ gets $reach(a,d)$ from $TB(reach(a,X))$ with mgu $\{X/d\}$}}}
\put(2326,-811){\makebox(0,0)[lb]{\smash{\SetFigFont{9}{10.8}{\rmdefault}{\mddefault}{\updefault}$\qquad\qquad\qquad\qquad$ $N_{14}:$  $\leftarrow return((d))$}}}
\put(4201,-1036){\makebox(0,0)[lb]{\smash{\SetFigFont{8}{9.6}{\rmdefault}{\mddefault}{\updefault}Return $X=d$}}}
\put(3601,-1411){\makebox(0,0)[lb]{\smash{\SetFigFont{9}{10.8}{\rmdefault}{\mddefault}{\updefault}$N_{15}:$ $\Box$}}}
\end{picture}

\caption{The third successful TP-derivation with the third answer $X=d$.}\label{fig-reach5}
\end{figure}
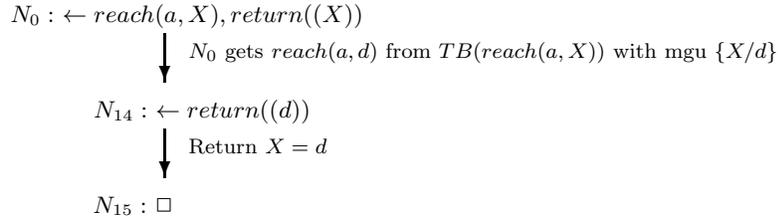 

\begin{figure}[p]
\setlength{\unitlength}{3947sp}%
\begingroup\makeatletter\ifx\SetFigFont\undefined%
\gdef\SetFigFont#1#2#3#4#5{%
  \reset@font\fontsize{#1}{#2pt}%
  \fontfamily{#3}\fontseries{#4}\fontshape{#5}%
  \selectfont}%
\fi\endgroup%
\begin{picture}(2400,1602)(1801,-1432)
\thicklines
\put(4051,-286){\vector( 0,-1){300}}
\put(4051,-886){\vector( 0,-1){300}}
\put(1951, 14){\makebox(0,0)[lb]{\smash{\SetFigFont{12}{14.4}{\rmdefault}{\mddefault}{\updefault}$\qquad$}}}
\put(1801,-211){\makebox(0,0)[lb]{\smash{\SetFigFont{9}{10.8}{\rmdefault}{\mddefault}{\updefault}$\qquad\qquad\qquad\qquad$ $N_0:$ $\leftarrow reach(a,X),return((X))$ }}}
\put(4201,-436){\makebox(0,0)[lb]{\smash{\SetFigFont{8}{9.6}{\rmdefault}{\mddefault}{\updefault}$N_0$ gets $reach(a,e)$ from $TB(reach(a,X))$ with mgu $\{X/e\}$}}}
\put(4201,-1036){\makebox(0,0)[lb]{\smash{\SetFigFont{8}{9.6}{\rmdefault}{\mddefault}{\updefault}Return $X=e$}}}
\put(3601,-1411){\makebox(0,0)[lb]{\smash{\SetFigFont{9}{10.8}{\rmdefault}{\mddefault}{\updefault}$N_{17}:$ $\Box$}}}
\put(2326,-811){\makebox(0,0)[lb]{\smash{\SetFigFont{9}{10.8}{\rmdefault}{\mddefault}{\updefault}$\qquad\qquad\qquad\qquad$ $N_{16}:$  $\leftarrow return((e))$}}}
\end{picture}

\caption{The fourth successful TP-derivation with the fourth answer $X=e$.}\label{fig-reach6}
\end{figure}
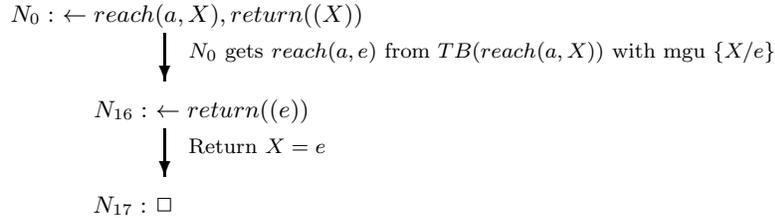
\pagebreak
The following example is also useful in illustrating 
TP-resolution.\footnote{This program is suggested by 
an anonymous referee.} To simplify the 
presentation, in the sequel, in depicting derivations
we omit subgoals like $memo\_look(.)$ and $return(.)$ unless they are required
to be explicitly present. 

\begin{example}
\label{p3}
{\em
Consider the logic program
\begin{tabbing}
\hspace{.2in} $P_2$: \= $p(a,b,c).$ \`$C_{p_1}$ \\
\> $p(X,Y,Z) \leftarrow p(Z,X,Y).$ \`$C_{p_2}$
\end{tabbing}
Choose $p$ as a tabled predicate. Let $G_0=\leftarrow p(X,Y,Z)$
be the top goal. The TP-tree of $P_2 \cup \{G_0\}$
consists of Figures \ref{fig-p31} and \ref{fig-p32},
which yields three answers, $p(a,b,c)$, $p(b,c,a)$ and $p(c,a,b)$.

\begin{figure}[htb]
\setlength{\unitlength}{3947sp}%
\begingroup\makeatletter\ifx\SetFigFont\undefined%
\gdef\SetFigFont#1#2#3#4#5{%
  \reset@font\fontsize{#1}{#2pt}%
  \fontfamily{#3}\fontseries{#4}\fontshape{#5}%
  \selectfont}%
\fi\endgroup%
\begin{picture}(5025,1545)(2101,-2485)
\thicklines
\put(4351,-1186){\vector(-3,-1){900}}
\put(4801,-1186){\vector( 3,-1){900}}
\put(5176,-1711){\vector(-3,-1){900}}
\put(6151,-1711){\vector( 3,-1){900}}
\put(5626,-1711){\vector( 0,-1){300}}
\put(5251,-2236){\makebox(0,0)[lb]{\smash{\SetFigFont{9}{10.8}{\rmdefault}{\mddefault}{\updefault}$N_4$:  $\Box$}}}
\put(3901,-2236){\makebox(0,0)[lb]{\smash{\SetFigFont{9}{10.8}{\rmdefault}{\mddefault}{\updefault}$N_3$:  $\Box$}}}
\put(2776,-1036){\makebox(0,0)[lb]{\smash{\SetFigFont{9}{10.8}{\rmdefault}{\mddefault}{\updefault}$\qquad\qquad\qquad\qquad$ $N_0$:  $\leftarrow p(X,Y,Z)$}}}
\put(2926,-1861){\makebox(0,0)[lb]{\smash{\SetFigFont{8}{9.6}{\rmdefault}{\mddefault}{\updefault}Add $p(a,b,c)$}}}
\put(3901,-2461){\makebox(0,0)[lb]{\smash{\SetFigFont{8}{9.6}{\rmdefault}{\mddefault}{\updefault}Add $p(b,c,a)$}}}
\put(5251,-2461){\makebox(0,0)[lb]{\smash{\SetFigFont{8}{9.6}{\rmdefault}{\mddefault}{\updefault}Add $p(c,a,b)$}}}
\put(3376,-1261){\makebox(0,0)[lb]{\smash{\SetFigFont{8}{9.6}{\rmdefault}{\mddefault}{\updefault}$C_{p_1}$}}}
\put(5326,-1261){\makebox(0,0)[lb]{\smash{\SetFigFont{8}{9.6}{\rmdefault}{\mddefault}{\updefault}$C_{p_2}$}}}
\put(5026,-1636){\makebox(0,0)[lb]{\smash{\SetFigFont{9}{10.8}{\rmdefault}{\mddefault}{\updefault}$N_2$:  $\leftarrow p(Z,X,Y)$}}}
\put(2101,-1636){\makebox(0,0)[lb]{\smash{\SetFigFont{9}{10.8}{\rmdefault}{\mddefault}{\updefault}$\qquad\qquad\qquad$ $N_1$:  $\Box$}}}
\put(4726,-2011){\makebox(0,0)[lb]{\smash{\SetFigFont{8}{9.6}{\rmdefault}{\mddefault}{\updefault}Get $p(a,b,c)$}}}
\put(7126,-2011){\makebox(0,0)[lb]{\smash{\SetFigFont{8}{9.6}{\rmdefault}{\mddefault}{\updefault}Get $p(c,a,b)$}}}
\put(5776,-2011){\makebox(0,0)[lb]{\smash{\SetFigFont{8}{9.6}{\rmdefault}{\mddefault}{\updefault}Get $p(b,c,a)$}}}
\put(6376,-2236){\makebox(0,0)[lb]{\smash{\SetFigFont{9}{10.8}{\rmdefault}{\mddefault}{\updefault}$N_5$:  $\leftarrow memo\_look(N_0,p(X,Y,Z),(a,b,c),\theta_0)$}}}
\end{picture}

\caption{TP-derivations of $P_2 \cup \{G_0\}$.
$\qquad\qquad\qquad\qquad$}\label{fig-p31}
\end{figure} 

\begin{figure}[htb]
\setlength{\unitlength}{3947sp}%
\begingroup\makeatletter\ifx\SetFigFont\undefined%
\gdef\SetFigFont#1#2#3#4#5{%
  \reset@font\fontsize{#1}{#2pt}%
  \fontfamily{#3}\fontseries{#4}\fontshape{#5}%
  \selectfont}%
\fi\endgroup%
\begin{picture}(5337,1545)(1726,-2560)
\thicklines
\put(5176,-1711){\vector(-3,-1){900}}
\put(6151,-1711){\vector( 3,-1){900}}
\put(5626,-1186){\vector( 0,-1){300}}
\put(5626,-1711){\vector( 0,-1){600}}
\put(5026,-1636){\makebox(0,0)[lb]{\smash{\SetFigFont{9}{10.8}{\rmdefault}{\mddefault}{\updefault}$N_6$:  $\leftarrow p(Z,X,Y)$}}}
\put(5101,-1111){\makebox(0,0)[lb]{\smash{\SetFigFont{9}{10.8}{\rmdefault}{\mddefault}{\updefault} $N_0$:  $\leftarrow p(X,Y,Z)$}}}
\put(4276,-2536){\makebox(0,0)[lb]{\smash{\SetFigFont{9}{10.8}{\rmdefault}{\mddefault}{\updefault}$N_8$:  $\leftarrow memo\_look(N_0,p(X,Y,Z),(c,a,b),\theta_0)$}}}
\put(5701,-1936){\makebox(0,0)[lb]{\smash{\SetFigFont{8}{9.6}{\rmdefault}{\mddefault}{\updefault}Get $p(b,c,a)$}}}
\put(5701,-1336){\makebox(0,0)[lb]{\smash{\SetFigFont{8}{9.6}{\rmdefault}{\mddefault}{\updefault}$C_{p_2}$}}}
\put(1726,-2236){\makebox(0,0)[lb]{\smash{\SetFigFont{9}{10.8}{\rmdefault}{\mddefault}{\updefault}$\quad$ $N_7$:  $\leftarrow memo\_look(N_0,p(X,Y,Z),(b,c,a),\theta_0)$}}}
\put(6526,-2236){\makebox(0,0)[lb]{\smash{\SetFigFont{9}{10.8}{\rmdefault}{\mddefault}{\updefault}$N_9$:  $\leftarrow memo\_look(N_0,p(X,Y,Z),(a,b,c),\theta_0)$}}}
\put(6751,-1861){\makebox(0,0)[lb]{\smash{\SetFigFont{8}{9.6}{\rmdefault}{\mddefault}{\updefault}Get $p(c,a,b)$}}}
\put(3826,-1861){\makebox(0,0)[lb]{\smash{\SetFigFont{8}{9.6}{\rmdefault}{\mddefault}{\updefault}Get $p(a,b,c)$}}}
\end{picture}

\caption{Answer iteration for $P_2 \cup \{G_0\}$.
$\qquad\qquad$}\label{fig-p32}
\end{figure}
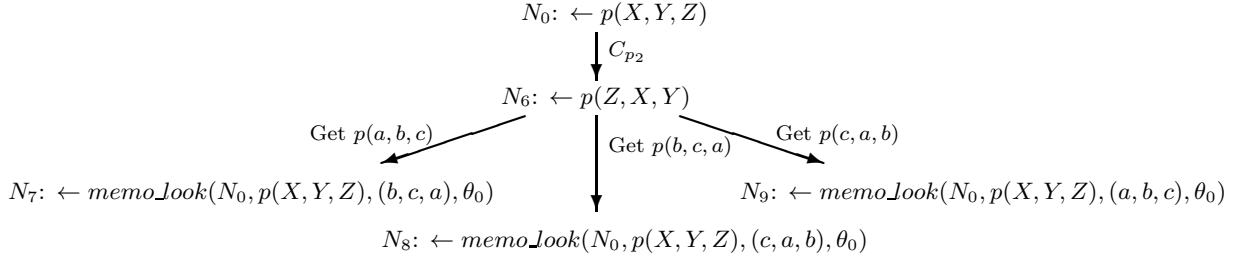

}
\end{example}

Note that in the above examples, no new answers are derived during
answer iteration (i.e. Algorithm 2 stops by the end of the first
iteration). We now give another example, which shows that
answer iteration is indispensable. 

\begin{example}
\label{lpe}
{\em
Consider the following logic program
\begin{tabbing}
\hspace{.2in} $P_3$: \= $p(X,Y) \leftarrow q(X,Y).$ \`$C_{p_1}$ \\
\> $q(X,Y) \leftarrow p(X,Z),t(Z,Y).$ \`$C_{q_1}$ \\
\> $q(a,b).$ \`$C_{q_2}$ \\
\> $t(b,c).$ \`$C_{t_1}$
\end{tabbing}

Choose $p$ and $q$ as tabled predicates and apply 
Algorithm 2 to the top goal $G_0=\leftarrow p(X,Y).$ 
After applying the clauses
$C_{p_1}$ and $C_{q_1}$, we generate the derivation 
shown in Figure \ref{fig-p21}. 
We see that a loop $L(N_0,N2)$ occurs. So we do not
use $C_{p_1}$ to expand $N_2$ because that would repeat the loop. Instead,
we try alternative clauses. Since there is no other clause in $P_3$ that 
is unifiable with $p(X,Z)$, we fail $N_2$ and backtrack to its 
parent node $N_1$, which leads to the derivation of Figure \ref{fig-p22}. 
Now, since there is no more clause available for $q(X,Y)$, we fail
$N_1$ and go back to $N_0$. Note that the flag $NEW$ has been set to $1$ 
because new answers, $q(a,b)$ and $p(a,b)$,
have been derived. Moreover, $C_{q_2}$ is {\em no longer available} to 
$q(X,Y)$, whereas both $C_{p_1}$ and $C_{q_1}$ are {\em still available}
because they are involved in a loop.

At $N_0$ answer iteration is performed. The first iteration is shown in
Figure \ref{fig-p23}, where two new answers, $q(a,c)$ and $p(a,c)$,
are derived. The second iteration will derive no new answers, so the
algorithm stops with the flag $COMP$ of $TB(p(X,Y))$ set to 1.

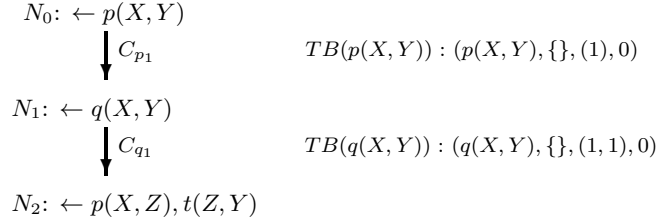
\begin{figure}[htb]
\setlength{\unitlength}{3947sp}%
\begingroup\makeatletter\ifx\SetFigFont\undefined%
\gdef\SetFigFont#1#2#3#4#5{%
  \reset@font\fontsize{#1}{#2pt}%
  \fontfamily{#3}\fontseries{#4}\fontshape{#5}%
  \selectfont}%
\fi\endgroup%
\begin{picture}(1950,1320)(3001,-2260)
\thicklines
\put(4876,-1711){\vector( 0,-1){300}}
\put(4876,-1111){\vector( 0,-1){300}}
\put(4951,-1261){\makebox(0,0)[lb]{\smash{\SetFigFont{8}{9.6}{\rmdefault}{\mddefault}{\updefault}$C_{p_1}$ $\qquad\qquad\qquad$ $TB(p(X,Y)):$ $(p(X,Y),\{\},(1),0)$}}}
\put(4951,-1861){\makebox(0,0)[lb]{\smash{\SetFigFont{8}{9.6}{\rmdefault}{\mddefault}{\updefault}$C_{q_1}$ $\qquad\qquad\qquad$ $TB(q(X,Y)):$ $(q(X,Y),\{\},(1,1),0)$}}}
\put(3076,-1036){\makebox(0,0)[lb]{\smash{\SetFigFont{9}{10.8}{\rmdefault}{\mddefault}{\updefault}$\qquad\qquad\qquad\qquad$ $N_0$:  $\leftarrow p(X,Y)$}}}
\put(3001,-1636){\makebox(0,0)[lb]{\smash{\SetFigFont{9}{10.8}{\rmdefault}{\mddefault}{\updefault}$\qquad\qquad\qquad\qquad$ $N_1$:  $\leftarrow q(X,Y)$}}}
\put(3001,-2236){\makebox(0,0)[lb]{\smash{\SetFigFont{9}{10.8}{\rmdefault}{\mddefault}{\updefault}$\qquad\qquad\qquad\qquad$ $N_2$:  $\leftarrow p(X,Z),t(Z,Y)$}}}
\end{picture}

\caption{A TP-derivation where a loop occurs.
$\qquad\qquad\qquad\qquad\qquad\qquad\qquad\qquad$}\label{fig-p21}
\end{figure}

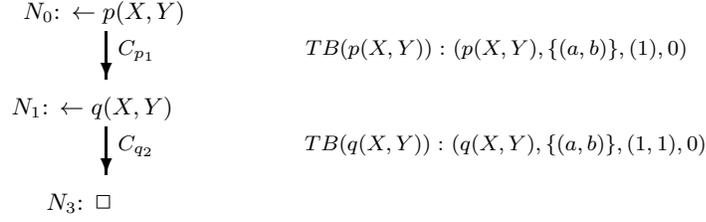
\begin{figure}[htb]
\setlength{\unitlength}{3947sp}%
\begingroup\makeatletter\ifx\SetFigFont\undefined%
\gdef\SetFigFont#1#2#3#4#5{%
  \reset@font\fontsize{#1}{#2pt}%
  \fontfamily{#3}\fontseries{#4}\fontshape{#5}%
  \selectfont}%
\fi\endgroup%
\begin{picture}(1950,1317)(3001,-2257)
\thicklines
\put(4876,-1711){\vector( 0,-1){300}}
\put(4876,-1111){\vector( 0,-1){300}}
\put(4951,-1261){\makebox(0,0)[lb]{\smash{\SetFigFont{8}{9.6}{\rmdefault}{\mddefault}{\updefault}$C_{p_1}$ $\qquad\qquad\qquad$ $TB(p(X,Y)):$ $(p(X,Y),\{(a,b)\},(1),0)$}}}
\put(4951,-1861){\makebox(0,0)[lb]{\smash{\SetFigFont{8}{9.6}{\rmdefault}{\mddefault}{\updefault}$C_{q_2}$ $\qquad\qquad\qquad$ $TB(q(X,Y)):$ $(q(X,Y),\{(a,b)\},(1,1),0)$}}}
\put(3076,-1036){\makebox(0,0)[lb]{\smash{\SetFigFont{9}{10.8}{\rmdefault}{\mddefault}{\updefault}$\qquad\qquad\qquad\qquad$ $N_0$:  $\leftarrow p(X,Y)$}}}
\put(3001,-1636){\makebox(0,0)[lb]{\smash{\SetFigFont{9}{10.8}{\rmdefault}{\mddefault}{\updefault}$\qquad\qquad\qquad\qquad$ $N_1$:  $\leftarrow q(X,Y)$}}}
\put(4501,-2236){\makebox(0,0)[lb]{\smash{\SetFigFont{9}{10.8}{\rmdefault}{\mddefault}{\updefault}$N_3$:  $\Box$}}}
\end{picture}

\caption{A successful TP-derivation.
$\qquad\qquad\qquad\qquad\qquad\qquad\qquad\qquad\qquad$}\label{fig-p22}
\end{figure}

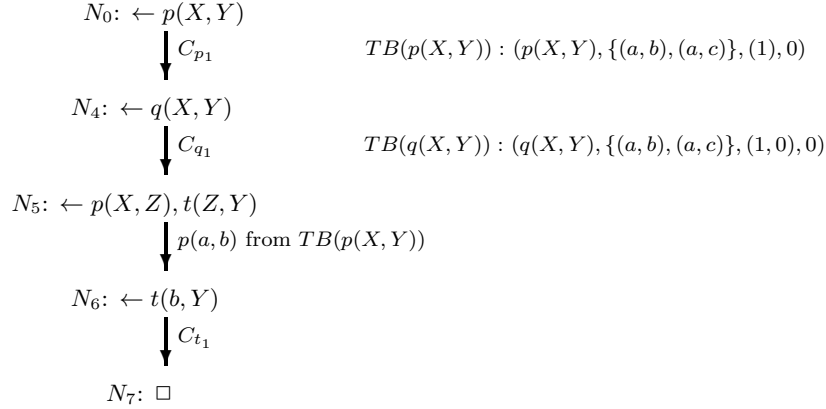
\begin{figure}[htb]
\setlength{\unitlength}{3947sp}%
\begingroup\makeatletter\ifx\SetFigFont\undefined%
\gdef\SetFigFont#1#2#3#4#5{%
  \reset@font\fontsize{#1}{#2pt}%
  \fontfamily{#3}\fontseries{#4}\fontshape{#5}%
  \selectfont}%
\fi\endgroup%
\begin{picture}(2325,2517)(2626,-3457)
\thicklines
\put(4876,-1711){\vector( 0,-1){300}}
\put(4876,-1111){\vector( 0,-1){300}}
\put(4876,-2311){\vector( 0,-1){300}}
\put(4876,-2911){\vector( 0,-1){300}}
\put(4951,-1261){\makebox(0,0)[lb]{\smash{\SetFigFont{8}{9.6}{\rmdefault}{\mddefault}{\updefault}$C_{p_1}$ $\qquad\qquad\qquad$ $TB(p(X,Y)):$ $(p(X,Y),\{(a,b),(a,c)\},(1),0)$}}}
\put(4951,-1861){\makebox(0,0)[lb]{\smash{\SetFigFont{8}{9.6}{\rmdefault}{\mddefault}{\updefault}$C_{q_1}$ $\qquad\qquad\qquad$ $TB(q(X,Y)):$ $(q(X,Y),\{(a,b),(a,c)\},(1,0),0)$}}}
\put(3076,-1036){\makebox(0,0)[lb]{\smash{\SetFigFont{9}{10.8}{\rmdefault}{\mddefault}{\updefault}$\qquad\qquad\qquad\qquad$ $N_0$:  $\leftarrow p(X,Y)$}}}
\put(3001,-1636){\makebox(0,0)[lb]{\smash{\SetFigFont{9}{10.8}{\rmdefault}{\mddefault}{\updefault}$\qquad\qquad\qquad\qquad$ $N_4$:  $\leftarrow q(X,Y)$}}}
\put(2626,-2236){\makebox(0,0)[lb]{\smash{\SetFigFont{9}{10.8}{\rmdefault}{\mddefault}{\updefault}$\qquad\qquad\qquad\qquad$ $N_5$:  $\leftarrow p(X,Z),t(Z,Y)$}}}
\put(4951,-2461){\makebox(0,0)[lb]{\smash{\SetFigFont{8}{9.6}{\rmdefault}{\mddefault}{\updefault}$p(a,b)$ from $TB(p(X,Y))$}}}
\put(3001,-2836){\makebox(0,0)[lb]{\smash{\SetFigFont{9}{10.8}{\rmdefault}{\mddefault}{\updefault}$\qquad\qquad\qquad\qquad$ $N_6$:  $\leftarrow t(b,Y)$}}}
\put(4951,-3061){\makebox(0,0)[lb]{\smash{\SetFigFont{8}{9.6}{\rmdefault}{\mddefault}{\updefault}$C_{t_1}$}}}
\put(4501,-3436){\makebox(0,0)[lb]{\smash{\SetFigFont{9}{10.8}{\rmdefault}{\mddefault}{\updefault}$N_7$:  $\Box$}}}
\end{picture}

\caption{New answers derived via answer iteration.
$\qquad\qquad\qquad\qquad\qquad\qquad\qquad$}\label{fig-p23}
\end{figure}

}
\end{example}

\section{Characteristics of TP-Resolution}

In this section, we prove the termination of Algorithm 2 and
the soundness and completeness of TP-resolution. 
We also discuss the way to
deal with the cut operator in TP-resolution.

\subsection{Soundness and Completeness}

In order to guarantee termination of Algorithm 2, we 
restrict ourselves to logic programs with the 
bounded-term-size property. The following definition
is adapted from \cite{VG89}. 

\begin{definition}
\label{bounded-term-size}
{\em
A logic program $P$ has the {\em bounded-term-size} property if there is a function
$f(n)$ such that whenever a top goal $G_0$ has no argument whose term size exceeds
$n$, then no subgoal in the TP-tree $TP_{G_0}$ and no answer tuple
in any table have an argument whose term size exceeds $f(n)$.
}
\end{definition} 

Obviously, all function-free logic programs have the 
bounded-term-size property.

\begin{theorem}[Termination]
\label{term}
Let $P$ be a logic program with 
the bounded-term-size property and $G_0$ a top goal.
Algorithm 2 terminates with a finite TP-tree $TP_{G_0}$.
\end{theorem}

The following lemma is required to prove this theorem.

\begin{lemma}
\label{lem2}
Let $G_i$ and $G_k$ be two goals in a TP-derivation
of $P \cup \{G_0\}$ and $A_i$ and $A_k$ be the left-most subgoals of $G_i$ and $G_k$,
respectively. If $A_i$ is an ancestor variant subgoal 
of $A_k$ then $A_i$ is a tabled subgoal.
\end{lemma}

\noindent {\bf Proof.}  Let $A_i=p(.)$. By Definitions \ref{alist}
and \ref{loop}, $A_i$ being an ancestor variant subgoal 
of $A_k$ implies that there is a cycle of the form 
$p \rightarrow ... \rightarrow p$ in the dependency graph
$G_P$. So $p$ is a tabled predicate and 
thus $A_i$ is a tabled subgoal. $\Box$\\[.15in]
{\bf Proof of Theorem \ref{term}.} 
Assume, on the contrary, that 
Algorithm 2 does not terminate.
Then it generates an infinite TP-tree.
This can occur only in two cases: (1) it memorizes infinitely
many (new) answers in tables, so we do backtracking 
at some nodes infinite times;
and (2) it traps into an infinite derivation. We first show that
the first case is not possible. Since $P$ has
the bounded-term-size property, all tabled facts have finite term
size. Then, in view of the fact that any logic program has only a 
finite number of predicate, function and constant
symbols, all tabled facts having finite term
size implies that any table has only a finite number of 
tabled facts.

We now assume the second case. Since $P$ has 
the bounded-term-size property
and contains only a finite number of clauses, any
infinite derivation must contain an infinite
loop, i.e. an infinite set of
subgoals, $A_0, A_1, ..., A_k, ...,$ 
such that for any $i \geq 0$, $A_i$ is both an
ancestor subgoal and a variant of $A_{i+1}$. 
This means that all the $A_i$s are tabled subgoals
(Lemma \ref{lem2}). However, from Cases 4 and 4.3 of Algorithm 2 
we see that such a set of subgoals
will never be generated
unless $P$ contains an infinite set of clauses whose
heads are unifiable with the $A_i$s, 
a contradiction.  $\Box$

To simplify the proof of soundness and completeness,
we assume, in the sequel, that all predicates are tabled predicates.
 
\begin{theorem}[Soundness and Completeness]
\label{soco}
Let $P$ be a logic program with the bounded-term-size
property and $G_0 = \leftarrow A_1,...,A_m$ a top goal
with the list $\vec{Y}$ of variables. Let $TP_{G_0}$ be the TP-tree of
$P \cup \{G_0\}$ and $ST_{G_0}$ the SLD-tree of $P \cup \{G_0\}$
via the left-most computation rule. Then
$TP_{G_0}$ and $ST_{G_0}$ have the same set of
answers to $G_0$.
\end{theorem}

{\bf Proof.} (Soundness) By the PMF mode, 
each tabled fact is an intermediate answer of some subgoal
(called {\em sub-refutation} in \cite{TS86})
in an SLD-derivation in $ST_{G_0}$. Since
the answer $\vec{I}$ returned at any success leaf in $TP_{G_0}$ is
an instance of $\vec{Y}$ such that each $A_i\vec{Y}/\vec{I}$ is 
an instance of a tabled fact, $\vec{Y}/\vec{I}$ must be the
answer substitution of some successful SLD-derivation in
$ST_{G_0}$.

(Completeness)  Algorithm 2 works in the same way as Algorithm 1 (i.e. it
expands and backtracks on nodes in the same way as Algorithm 1) except (1) it
is based on the PMF mode, (2) after finishing backtracking for
answers of a subgoal $A_f=q(.)$ through the application of a clause
$C_{q_j}$, the status of $C_{q_j}$ w.r.t. $A_f$ will be set to
{\em ``no longer available''} (see Case B3), and (3) loops are handled
by skipping repeated clauses and doing answer iteration. 
Since the PMF mode preserves the answers of SLD-resolution and point (2) is only 
for the purpose of avoiding redundant computations (i.e. 
when variant subgoals of $A_f$ later occur,
they will directly use the tabled answers instead of recomputing them by
applying $C_{q_j}$), it suffices to prove that point (3) does not lose
any answers to $G_0$. 

\begin{figure}[htb]

\setlength{\unitlength}{3947sp}%
\begingroup\makeatletter\ifx\SetFigFont\undefined%
\gdef\SetFigFont#1#2#3#4#5{%
  \reset@font\fontsize{#1}{#2pt}%
  \fontfamily{#3}\fontseries{#4}\fontshape{#5}%
  \selectfont}%
\fi\endgroup%
\begin{picture}(3000,4080)(301,-4036)
\thinlines
\put(3226,-1186){\vector( 0,-1){225}}
\put(3226,-586){\vector( 0,-1){225}}
\put(3226,-1861){\vector( 0,-1){225}}
\thicklines
\multiput(3226,-166)(0.00000,-60.00000){3}{\makebox(6.6667,10.0000){\SetFigFont{10}{12}{\rmdefault}{\mddefault}{\updefault}.}}
\multiput(3226,-1441)(0.00000,-60.00000){3}{\makebox(6.6667,10.0000){\SetFigFont{10}{12}{\rmdefault}{\mddefault}{\updefault}.}}
\multiput(3226,-841)(0.00000,-60.00000){3}{\makebox(6.6667,10.0000){\SetFigFont{10}{12}{\rmdefault}{\mddefault}{\updefault}.}}
\multiput(3226,-2116)(0.00000,-60.00000){3}{\makebox(6.6667,10.0000){\SetFigFont{10}{12}{\rmdefault}{\mddefault}{\updefault}.}}
\multiput(3226,-2641)(0.00000,-60.00000){3}{\makebox(6.6667,10.0000){\SetFigFont{10}{12}{\rmdefault}{\mddefault}{\updefault}.}}
\multiput(3226,-3166)(0.00000,-60.00000){3}{\makebox(6.6667,10.0000){\SetFigFont{10}{12}{\rmdefault}{\mddefault}{\updefault}.}}
\multiput(3226,-3691)(0.00000,-60.00000){3}{\makebox(6.6667,10.0000){\SetFigFont{10}{12}{\rmdefault}{\mddefault}{\updefault}.}}
\put(2926,-61){\makebox(0,0)[lb]{\smash{\SetFigFont{9}{10.8}{\rmdefault}{\mddefault}{\updefault}$N_0:$ $G_0$}}}
\put(3301,-1336){\makebox(0,0)[lb]{\smash{\SetFigFont{7}{8.4}{\rmdefault}{\mddefault}{\updefault}$C_{p_j}$}}}
\put(1576,-511){\makebox(0,0)[lb]{\smash{\SetFigFont{9}{10.8}{\rmdefault}{\mddefault}{\updefault}$\qquad\qquad\qquad\quad$ $N_{l_0}:$ $\leftarrow p(\vec{X_0}),B_0$}}}
\put(3301,-736){\makebox(0,0)[lb]{\smash{\SetFigFont{7}{8.4}{\rmdefault}{\mddefault}{\updefault}$C_{p_j}$}}}
\put(3301,-2011){\makebox(0,0)[lb]{\smash{\SetFigFont{7}{8.4}{\rmdefault}{\mddefault}{\updefault}$C_{p_j}$}}}
\put(301,-1786){\makebox(0,0)[lb]{\smash{\SetFigFont{9}{10.8}{\rmdefault}{\mddefault}{\updefault}$\qquad\qquad\qquad\quad$ $N_{l_m}:$ $\leftarrow p(\vec{X}_m),B_m,B_{m-1}\theta_{m-1},...,B_1\theta_{m-1}...\theta_1,B_0\theta_{m-1}...\theta_0$}}}
\put(301,-2461){\makebox(0,0)[lb]{\smash{\SetFigFont{9}{10.8}{\rmdefault}{\mddefault}{\updefault}$\qquad\qquad\qquad\quad$ $N_{x_m}:$ $\leftarrow B_m\gamma_m,B_{m-1}\gamma_m\theta_{m-1},...,B_1\gamma_m\theta_{m-1}...\theta_1,B_0\gamma_m\theta_{m-1}...\theta_0$}}}
\put(526,-2986){\makebox(0,0)[lb]{\smash{\SetFigFont{9}{10.8}{\rmdefault}{\mddefault}{\updefault}$\qquad\qquad\qquad\quad$ $N_{x_1}:$ $\leftarrow B_1\gamma_m...\gamma_1\theta_{m-1}...\theta_1,B_0\gamma_m...\gamma_1\theta_{m-1}...\theta_0$}}}
\put(1351,-1111){\makebox(0,0)[lb]{\smash{\SetFigFont{9}{10.8}{\rmdefault}{\mddefault}{\updefault}$\qquad\qquad\qquad\quad$ $N_{l_1}:$ $\leftarrow p(\vec{X_1}),B_1,B_0\theta_0$}}}
\put(1201,-3511){\makebox(0,0)[lb]{\smash{\SetFigFont{9}{10.8}{\rmdefault}{\mddefault}{\updefault}$\qquad\qquad\qquad\quad$ $N_{x_0}:$ $\leftarrow B_0\gamma_m...\gamma_0\theta_{m-1}...\theta_0$}}}
\put(2851,-4036){\makebox(0,0)[lb]{\smash{\SetFigFont{9}{10.8}{\rmdefault}{\mddefault}{\updefault}$N_t:$ $\Box$}}}
\end{picture}

\caption{An SLD-derivation with loops. $\qquad\qquad$}\label{fig-proof}
\end{figure}
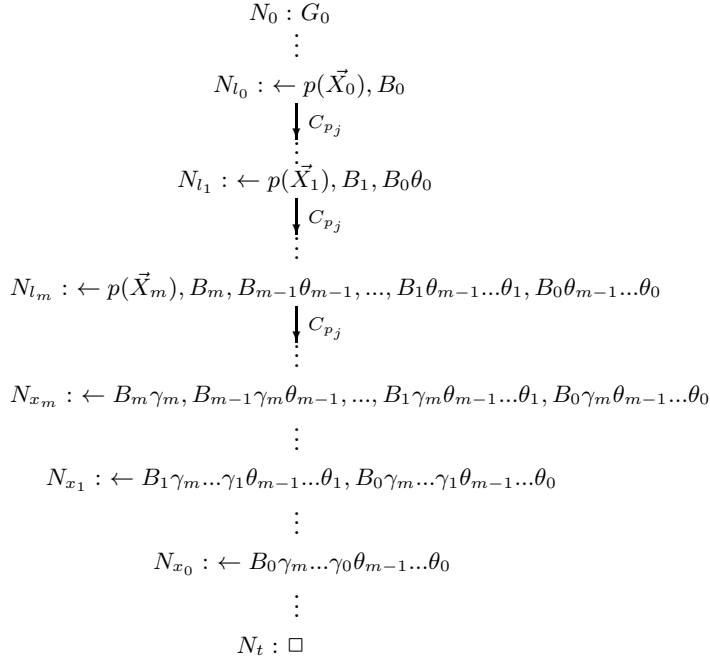

Let $SD$ be an arbitrary successful SLD-derivation in $ST_{G_0}$ with loops
as shown in Figure \ref{fig-proof}, 
where $m>0$, $N_{l_0}$ is an iteration node and for any $0\leq i <m$
$p(\vec{X_i})$ is an ancestor variant 
subgoal of $p(\vec{X}_{i+1})$. Note that the SLD-derivation starts looping
at $N_{l_1}$ by applying $C_{p_j}$ to $p(\vec{X_1})$. However,
Algorithm 2 will handle such loops by skipping $C_{p_j}$ at $N_{l_1}$
and doing answer iteration at $N_{l_0}$. Before showing that 
no answers to $p(\vec{X_0})$ will be lost using the skipping-iterating technique, 
we further explain the structure of the loops in $SD$ as follows.
\begin{enumerate}
\item[(1)]
For $0\leq i <m$ from $N_{l_i}$ to $N_{l_{i+1}}$ 
the proof of $p(\vec{X_i})$ reduces to 
the proof of $(p(\vec{X}_{i+1}),B_{i+1})$ with
a substitution $\theta_i$ for $p(\vec{X_i})$, where
each $B_k$ $(0\leq k \leq m)$ is a set of subgoals. 

\item[(2)]
The sub-refutation between $N_{l_m}$ and $N_{x_m}$ 
contains no loops and yields an answer
$p(\vec{X}_m)\gamma_m$ to $p(\vec{X}_m)$. The answer substitution
$\gamma_m$ for $p(\vec{X}_m)$ is then applied to the remaining
subgoals of $N_{l_m}$ (see node $N_{x_m}$), which leads to an answer
$p(\vec{X}_{m-1})\gamma_m\gamma_{m-1}\theta_{m-1}$ to $p(\vec{X}_{m-1})$. 
Such process continues recursively until an answer
$p(\vec{X_0})\gamma_m...\gamma_0\theta_{m-1}...\theta_0$ to $p(\vec{X_0})$
is produced at $N_{x_0}$.
\end{enumerate} 

We now prove that a variant of the answer
$p(\vec{X_0})\gamma_m...\gamma_0\theta_{m-1}...\theta_0$ 
to $p(\vec{X_0})$ will be produced by Algorithm 2 by means of answer iteration.

Since $p(\vec{X}_0)$ and $p(\vec{X}_m)$ are variants, via backtracking
from $N_{l_1}$ up to $N_{l_0}$ a variant of the sub-refutation between
$N_{l_m}$ and $N_{x_m}$ can be generated, which starts from $N_{l_0}$
via $C_{p_j}$. This means that a variant of the answer
$p(\vec{X}_m)\gamma_m$ to $p(\vec{X}_m)$ can be derived via backtracking
from $N_{l_1}$ up to $N_{l_0}$, independently of the sub-derivation
below $N_{l_1}$. 

Let us do backtracking from $N_{l_1}$ up to $N_{l_0}$ and store all
intermediate answers in their tables. So $p(\vec{X}_m)\gamma_m$
is in $TB(p(\vec{X}_0))$. Now we regenerate the loop $L(N_{l_0},N_{l_1})$
(the first iteration).

Since $p(\vec{X}_0)$ and $p(\vec{X}_{m-1})$ are variants, a variant of 
the sub-refutation between $N_{l_{m-1}}$ and $N_{x_{m-1}}$, where
the sub-refutation between $N_{l_m}$ and $N_{x_m}$ is replaced by directly 
using the answer $p(\vec{X}_m)\gamma_m$, can be generated via backtracking
from $N_{l_1}$ up to $N_{l_0}$. That is, a variant of the answer
$p(\vec{X}_{m-1})\gamma_m\gamma_{m-1}\theta_{m-1}$ to $p(\vec{X}_{m-1})$
can also be derived via backtracking from $N_{l_1}$ up to $N_{l_0}$ when
the tabled answer $p(\vec{X}_m)\gamma_m$ is used.
So we do the backtracking, store $p(\vec{X}_{m-1})\gamma_m\gamma_{m-1}\theta_{m-1}$
in $TB(p(\vec{X}_0))$, and then
regenerate the loop $L(N_{l_0},N_{l_1})$ (the second iteration).

Continue the above process recursively. 
After (at most) $m$ iterations, a variant of the answer
$p(\vec{X_0})\gamma_m...\gamma_0\theta_{m-1}...\theta_0$ to $p(\vec{X_0})$
will be derived and stored in $TB(p(\vec{X}_0))$.

The above arguments show that although the branch below $N_{l_1}$ via $C_{p_j}$
is skipped by Algorithm 2, by means of answer iteration along with tabling
no answers will be lost to $p(\vec{X_0})$. Therefore, when  
a fixpoint is reached at $N_{l_0}$, which means no new answers to $p(\vec{X}_0)$ 
can be derived via iterations, all answers of $p(\vec{X}_0)$ must be exhausted and stored 
in $TB(p(\vec{X}_0))$ (in such a case, the flag $TB(p(\vec{X}_0)) \rightarrow COMP$
is set to 1). We now prove that the fixpoint can be reached in finite time
even if $m\rightarrow \infty$.

Let $m\rightarrow \infty$. Then $SD$ contains infinite loops. 
Since $P$ has the bounded-term-size property and only
a finite number of clauses, we have only a finite number of subgoals and
any subgoal has only a finite number of answers (up to variable 
renaming). Let $N$ be the number of all answers of all subgoals.
Since before the fixpoint is reached, in each iteration at $N_{l_0}$ 
at least one new answer to some subgoal will be derived, the fixpoint
will be reached after at most $N$ iterations.

To sum up, Algorithm 2 traverses $ST_{G_0}$ as follows:
For any SLD-derivation $SD$ in $ST_{G_0}$, if it has no loops
Algorithm 2 will generate it based on the PMF mode while removing redundant
application of clauses; otherwise, Algorithm 2 will derive the answers
of subgoals involved in the loops by means of answer iteration. In either
case, Algorithm 2 terminates and preserves the answers of SLD-resolution.
As a result, if $SD$ is successful with an answer to $G_0$, there must
be a successful TP-derivation in $TP_{G_0}$ with the same answer 
(up to variable renaming). $\Box$

\subsection{Dealing with Cuts}
The cut operator, !, is very popular in Prolog programming. It
basically serves two purposes. One is to simulate
the {\em if-then-else} statement, which is one of
the key flow control statements in procedural languages. For example,
in order to realize the statement 
{\em if-$A$-then-$B$-else-$C$}, we define the following:

\begin{tabbing}
$\qquad\quad $ \= $H \leftarrow A, !, B$. \\
\> $H \leftarrow C$.
\end{tabbing}

\noindent The other, perhaps more important, purpose of using cuts is 
to prune the search space by aborting further exploration of 
some remaining branches, which may lead to significant computational
savings. For instance, the following clauses

\begin{tabbing}
$\qquad\quad $ \= $p(\vec{X}) \leftarrow A_1, ..., A_m, !$. \\
\> {\em $C_{p(.)}$: the remaining clauses defining $p(.)$}.
\end{tabbing}

\noindent achieve the effect that for any $\vec{X}$ whenever
$A_1, ..., A_m$ is true with an mgu $\theta$, we return $p(\vec{X})\theta$
and stop searching the remaining space
(via backtracking on the $A_i$s and using the remaining clauses
$C_{p(.)}$) for any additional answers of $p(\vec{X})$. 

The cut operator requires a strictly sequential strategy 
$-$ Prolog-strategy for the selection of goals, subgoals 
and program clauses. TP-strategy is an enhancement
of Prolog-strategy with the following two policies for dealing with
tabled facts (see Definition \ref{tps}): 
Table-first when both tabled facts and
program clauses are available and Top-down for the selection
of tabled facts. Since new answers are always appended to 
the end of tables, by the PMF mode, such an enhancement does
not affect the original sequentiality
of Prolog-strategy. That is, TP-strategy supports the cut
operator as well. 

Before enhancing Algorithm 2 with mechanisms for handling
cuts, we recall the operational semantics of cuts.

\begin{definition}
\label{cut}
{\em
Let $P$ be a logic program that contains the following clauses
with a head $p(.)$:
\begin{tabbing}
\hspace{.2in} \= $p(.)$ \= $\leftarrow ....$ \`$C_{p_1}$ \\
\>\> \vdots \\
\> $p(\vec{Y}) \leftarrow B_1,...,B_m,!,B_{m+2},....,B_{m+k}$ \`$C_{p_i}$ \\
\>\> \vdots\\
\> $p(.) \leftarrow ....$ \`$C_{p_n}$
\end{tabbing}
Let $p(\vec{X})$ be a subgoal such that $p(\vec{X})\theta=p(\vec{Y})\theta$. 
The semantics of ! in $C_{p_i}$ is defined as follows:
During top-down evaluation of $p(\vec{X})$, by the left-most
computation rule whenever $(B_1,...,B_m)\theta$ succeeds
with an mgu $\theta_1$, all the remaining answers to the subgoal $p(\vec{X})$
are obtained by computing 
$(B_{m+2},....,B_{m+k})\theta\theta_1$, with the backtracking on 
the $B_j$s ($1 \leq j \leq m$) and
the remaining clauses $C_{p_j}$s ($i < j \leq n$) ignored. In other
words, we force two {\em skips} when backtracking on the cut:
the skip of all $B_j$s ($1 \leq j \leq m$) and the skip of all
$C_{p_j}$s ($i < j \leq n$). 
}
\end{definition}

It is quite easy to realize cuts in TP-resolution. 
Let $N_h$ be a node labeled by a goal
 
$\quad G_h=\leftarrow p(\vec{X}),...$

\noindent and the clauses for $p(.)$ be as in Definition \ref{cut}. Let  

$\quad G_{h+1}=\leftarrow (B_1,...,B_m)\theta,!,(B_{m+2},...,B_{m+k})\theta,...$

\noindent be the resolvent of $G_h$ and $C_{p_i}$.
When evaluated as a subgoal for forward node expansion, 
$!$ is unconditionally true. However, during backtracking, by Definition \ref{cut}
it will skip all $B_j$s by directly jumping back to the node $N_h$.
In order to formalize such a jump, we attach to the subgoal $!$ 
a node name $N_h$ as a directive for
backtracking. That is, we create a subgoal $!(N_h)$, instead of
$!$, in the resolvent $G_{h+1}$. 

Then cuts are realized in TP-resolution
simply by adding to Algorithm 2, before Case 1 in point 2, the case
\begin{itemize}
\item
Case 0: $A_1$ is $!(N_h)$. Set $G_{i+1}=\leftarrow A_2,...,A_m$
and $N_h \rightarrow node\_SUSP=0$, and goto 2.
\end{itemize}
\noindent and, before Case B1 in point 3, the case
\begin{itemize}
\item
Case B0: $A_f$ is $!(N_h)$. Let $A_h=p(\vec{X})$ be 
the left-most subgoal at $N_h$ and $C_{p_i}$ be the clause that
is being used by $A_h$. If $A_h$ is a non-tabled subgoal then goto
3 with $N=N_h$. Otherwise, if $N_h \rightarrow node\_SUSP=0$
then set $TB(A_h) \rightarrow clause\_status[j]=0$ for all $j \geq i$;
else set $N_h \rightarrow node\_SUSP=0$ and 
$N_h \rightarrow clause\_ptr=null$.
Goto $2$ with $N_h$ as the latest registered node.
\end{itemize}

For Case 0, since $!$ is unconditionally true, $G_{i+1}=\leftarrow A_2,...,A_m$.
For Case B0, we do backtracking on the subgoal $!(N_h)$ at node $N_f$.
By Definition \ref{cut},
we will skip all nodes used for evaluating $(B_1,...,B_m)\theta$
and then skip all clauses $C_{p_j}$s with $j>i$. 
The first skip is done by jumping from $N_f$ back to $N_h$.
If $p(\vec{X})$ at $N_h$ is a non-tabled subgoal, the second skip
is done by failing the subgoal via backtracking. Otherwise,
we consider two cases. 

\begin{enumerate}
\item
Assume $N_h \rightarrow node\_SUSP=0$.
This means the evaluation of $(B_{m+2},....,B_{m+k})\theta\theta_1$
did not encounter any loop that goes through $N_h$ via $C_{p_i}$,
so that all answers of $(B_{m+2},....,B_{m+k})\theta\theta_1$
must have been exhausted via backtracking. Thus there will be no new answers
of $p(\vec{X})$ that can be derived
by applying the clauses $C_{p_j}$s $(j\geq i)$. Therefore, in
this case the second skip is achieved by changing the status of the $C_{p_j}$s
in $TB(p(\vec{X}))$ to {\em ``no longer available''}.

\item
Assume $N_h \rightarrow node\_SUSP=1$. Since the flag
$N_h \rightarrow node\_SUSP$ is initialized to 0 after
the evaluation of $(B_1,...,B_m)\theta$ (see Case 0), 
$N_h \rightarrow node\_SUSP=1$ means
that the evaluation of $(B_{m+2},....,B_{m+k})\theta\theta_1$
encountered loops that go through $N_h$ via $C_{p_i}$. So answer iteration
is required to exhaust the answers of $(B_{m+2},....,B_{m+k})\theta\theta_1$. 
Hence, in this case the
second skip is done simply by clearing the pointer $N_h \rightarrow clause\_ptr$,
so that no more clauses will be available to $p(\vec{X})$ at $N_h$.
\end{enumerate}

\begin{example}
\label{exp3}
{\em
Consider the following logic program: 
\begin{tabbing}
\hspace{.2in} $P_4$: \= $p(X,Y) \leftarrow p(X,Z),t(Z,Y).$ \`$C_{p_1}$ \\
\> $p(X,Y) \leftarrow p(X,Y),!.$ \`$C_{p_2}$ \\
\> $p(a,b).$ \`$C_{p_3}$ \\
\> $p(f,g).$ \`$C_{p_4}$ \\
\> $t(b,c).$ \`$C_{t_1}$
\end{tabbing}
Choose $p$ as a tabled predicate. Let $G_0=\leftarrow p(X,Y)$ 
be the top goal. By applying $C_{p_1}$ to the
root $N_0$ we generate $N_1$, where the first loop $L(N_0,N_1)$ occurs
(see Figure \ref{fig-cut1}).
Then $C_{p_2}$ is applied, which yields the second loop $L(N_1,N_2)$.
Since $C_{p_2}$ is being used by $N_1$, $C_{p_3}$ is used
to expand $N_2$, which gives the first
tabled fact $p(a,b)$. At $N_3$, the cut succeeds unconditionally, which leads to
$N_4$. Then $C_{t_1}$ is applied, giving the first success leaf $N_5$
with the second tabled fact $p(a,c)$ added to $TB(p(X,Y))$.

\begin{figure}[htb]
\setlength{\unitlength}{3947sp}%
\begingroup\makeatletter\ifx\SetFigFont\undefined%
\gdef\SetFigFont#1#2#3#4#5{%
  \reset@font\fontsize{#1}{#2pt}%
  \fontfamily{#3}\fontseries{#4}\fontshape{#5}%
  \selectfont}%
\fi\endgroup%
\begin{picture}(4950,2817)(901,-3832)
\thicklines
\put(4351,-1186){\vector(-3,-1){900}}
\put(4801,-1186){\vector( 3,-1){900}}
\put(3226,-1711){\vector(-3,-1){900}}
\put(2176,-2311){\vector( 0,-1){300}}
\put(2176,-2836){\vector( 0,-1){300}}
\put(2176,-3361){\vector( 0,-1){300}}
\put(3451,-1711){\vector( 3,-1){900}}
\put(3376,-1261){\makebox(0,0)[lb]{\smash{\SetFigFont{8}{9.6}{\rmdefault}{\mddefault}{\updefault}$C_{p_1}$}}}
\put(2551,-1636){\makebox(0,0)[lb]{\smash{\SetFigFont{9}{10.8}{\rmdefault}{\mddefault}{\updefault}$N_1$:  $\leftarrow p(X,Z),t(Z,Y)$}}}
\put(5176,-1261){\makebox(0,0)[lb]{\smash{\SetFigFont{8}{9.6}{\rmdefault}{\mddefault}{\updefault}Get $p(a,c)$ from $TB(p(X,Y))$}}}
\put(5401,-1636){\makebox(0,0)[lb]{\smash{\SetFigFont{9}{10.8}{\rmdefault}{\mddefault}{\updefault}$N_7$:  $\Box$}}}
\put(3976,-1111){\makebox(0,0)[lb]{\smash{\SetFigFont{9}{10.8}{\rmdefault}{\mddefault}{\updefault}$N_0$:  $\leftarrow p(X,Y)$}}}
\put(5851,-1636){\makebox(0,0)[lb]{\smash{\SetFigFont{8}{9.6}{\rmdefault}{\mddefault}{\updefault}Return $X=a$ and $Y=c$}}}
\put(2251,-2461){\makebox(0,0)[lb]{\smash{\SetFigFont{8}{9.6}{\rmdefault}{\mddefault}{\updefault}Add $p(a,b)$ to $TB(p(X,Y))$}}}
\put(1276,-2761){\makebox(0,0)[lb]{\smash{\SetFigFont{9}{10.8}{\rmdefault}{\mddefault}{\updefault}$N_3$:  $\leftarrow !(N_1),t(b,Y)$}}}
\put(2251,-3511){\makebox(0,0)[lb]{\smash{\SetFigFont{8}{9.6}{\rmdefault}{\mddefault}{\updefault}Add $p(a,c)$ to $TB(p(X,Y))$}}}
\put(1801,-3811){\makebox(0,0)[lb]{\smash{\SetFigFont{9}{10.8}{\rmdefault}{\mddefault}{\updefault}$N_5$:  $\Box$}}}
\put(1801,-2461){\makebox(0,0)[lb]{\smash{\SetFigFont{8}{9.6}{\rmdefault}{\mddefault}{\updefault}$C_{p_3}$}}}
\put(1801,-3511){\makebox(0,0)[lb]{\smash{\SetFigFont{8}{9.6}{\rmdefault}{\mddefault}{\updefault}$C_{t_1}$}}}
\put(1651,-3286){\makebox(0,0)[lb]{\smash{\SetFigFont{9}{10.8}{\rmdefault}{\mddefault}{\updefault}$N_4$:  $\leftarrow t(b,Y)$}}}
\put(2326,-3811){\makebox(0,0)[lb]{\smash{\SetFigFont{8}{9.6}{\rmdefault}{\mddefault}{\updefault}Return $X=a$ and $Y=b$}}}
\put(4051,-1861){\makebox(0,0)[lb]{\smash{\SetFigFont{8}{9.6}{\rmdefault}{\mddefault}{\updefault}Get $p(a,c)$ from $TB(p(X,Y))$}}}
\put(2101,-1861){\makebox(0,0)[lb]{\smash{\SetFigFont{8}{9.6}{\rmdefault}{\mddefault}{\updefault}$C_{p_2}$}}}
\put(3976,-2236){\makebox(0,0)[lb]{\smash{\SetFigFont{9}{10.8}{\rmdefault}{\mddefault}{\updefault}$N_6$:  $\leftarrow t(c,Y)$}}}
\put(901,-2236){\makebox(0,0)[lb]{\smash{\SetFigFont{9}{10.8}{\rmdefault}{\mddefault}{\updefault}$\qquad$ $N_2$:  $\leftarrow p(X,Z),!(N_1),t(Z,Y)$}}}
\end{picture}

\caption{TP-derivations with cuts.
$\qquad\qquad\qquad\qquad$}\label{fig-cut1}
\end{figure}
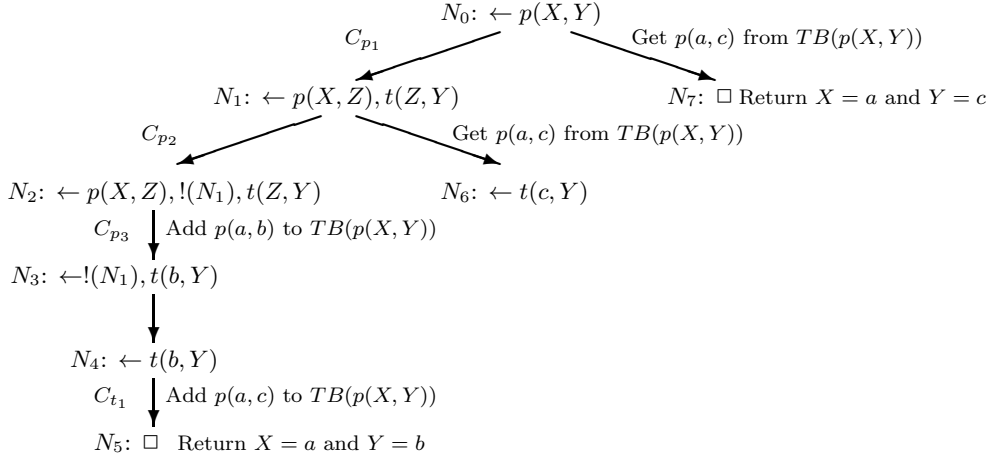

We backtrack to $N_4$ and then to $N_3$.
Due to the subgoal $!(N_1)$, we directly backtrack to $N_1$ (the first skip). 
The status of $C_{p_2}$, $C_{p_3}$ and $C_{p_4}$
in $TB(p(X,Y))$ is then changed to {\em ``no longer available''} (the second skip).
At $N_1$, the second tabled fact $p(a,c)$ is used, which yields
a failure leaf $N_6$. Next we go back to $N_0$, where the second 
tabled fact $p(a,c)$ is used, which gives the second success leaf $N_7$.
}
\end{example}

Similar extension can be made to Algorithm 1 to deal with cuts
in Prolog. By comparison of the two, we see that without loops,
cuts in TP-resolution achieve the same effect as in Prolog.
When there are loops, however, TP-resolution still reaches conclusions,
whereas Prolog will never stop.
The following representative example illustrates such a
difference.  

\begin{example}
\label{p5}
{\em The following two clauses 
\begin{tabbing}
\hspace{.2in} \= $not\_p(X) \leftarrow p(X),!,fail.$\`$C_{np_1}$ \\
\> $not\_p(X).$ \`$C_{np_2}$ 
\end{tabbing}
\noindent define the predicate $not\_p$ which says that
for any object $X$, $not\_p(X)$ succeeds if and only if
$p(X)$ fails. Let $G_0=\leftarrow not\_p(a)$ be 
the top goal and the programs $P_{5_i}$ be defined as follows.

\begin{enumerate}
\item
$P_{5_1}=\{C_{np_1},C_{np_2}\}$. As $p(a)$ fails, $C_{np_2}$ is applied, so that 
both Prolog and Algorithm 2 give an answer $YES$ to $G_0$.

\item
$P_{5_2}=\{C_{np_1},C_{np_2},p(a)\}$. As $p(a)$ succeeds, the cut 
$!$ in $C_{np_1}$ is executed. Since the subgoal $fail$ always
fails, the backtracking on $!$ skips $C_{np_2}$,
so that both Prolog and Algorithm 2 give an answer $NO$ to $G_0$.

\item
$P_{5_3}=\{C_{np_1},C_{np_2},p(X) \leftarrow p(X)\}$. 
Note that $p$ is a tabled predicate.
As Prolog goes into an infinite loop in proving the subgoal $p(a)$,
no answer to $G_0$ can be obtained. However, Algorithm 2
breaks the loop by deriving a negative answer to $p(a)$, so that
$C_{np_2}$ is applied, which leads to an answer $YES$ to $G_0$.
\end{enumerate}

}
\end{example}

As we mentioned earlier, cuts are used for two main purposes:
(1) simulate the {\em if-$A$-then-$B$-else-$C$} statement, i.e. treat
$B$ and $C$ to be two exclusive objects; (2) prune the search
space, i.e. force the two skips when backtracking on cuts (see 
Definition \ref{cut}). Since the second purpose exactly corresponds to
the operational semantics of cuts, it is achieved by
both Prolog and TP-resolution in any situations.
It turns out, however, that the first purpose cannot
be achieved in arbitrary situations. The following example
illustrates this.

\begin{example}
\label{p6}
{\em
Consider the following logic program: 
\begin{tabbing}
\hspace{.2in} $P_6$: \= $p(X) \leftarrow q(X),p(b),!,B$. \`$C_{p_1}$\\
\> $p(X) \leftarrow C$. \`$C_{p_2}$\\
\> $q(a).$ \`$C_{q_1}$ \\
\> $B.$ \`$C_{B_1}$ \\
\> $C.$ \`$C_{C_1}$
\end{tabbing}
It is easy to check that this program will generate no loops.
However, the two clauses $C_{p_1}$ and $C_{p_2}$ do not represent

{\em $\quad$ if $q(X)$ and $p(b)$ then $B$ else $C$} 

\noindent because evaluating $p(X)$ by Prolog/TP-resolution will lead to both
$C$ and $B$ being executed, which violates the intension 
that they are exclusive objects.
}
\end{example}

\begin{definition}
\label{cut0}
{\em Let $P$ be a program.
We say that the effect of {\em if-A-then-B-else-C} is achieved 
using clauses of the form
\begin{tabbing}
$\quad\quad$ \= $H \leftarrow A,!,B$.\\
\> $H \leftarrow C$.
\end{tabbing}
if when evaluating $H$ against $P$, either $B$ (i.e. when $A$ is true) 
or $C$ (i.e. when $A$ is false) but not both will be executed.
}
\end{definition}

Based on this criterion, we give
the following characterizations of the classes of programs
for which cuts are effectively handled by Prolog/TP-resolution
to achieve the effect of {\em if-A-then-B-else-C}.

\begin{theorem}
\label{cut1}
Let $P$ be a program with the bounded-term-size property. 
Let $A=A_1,...,A_m$, $B=B_1,...,B_n$ and
$C=C_1,...,C_q$. TP-resolution achieves the effect of 
if-$A$-then-$B$-else-$C$ using the following clauses in $P$
\begin{tabbing}
$\quad\quad$ \= $H \leftarrow A,!,B$.\\
\> $H \leftarrow C$.
\end{tabbing}
if and only if (1) if $A$ is true with the first answer substitution $\theta$
then the evaluation of $A$ for the first answer and the evaluation of $B\theta$
will not invoke $C$; (2) if $A$ is false
then the evaluation of $A$ and the evaluation of $C$ will not invoke $B$.
\end{theorem}

\noindent {\bf Proof.} $(\Longrightarrow)$ Straightforward. 

$(\Longleftarrow)$ Since TP-resolution always terminates, the truth
value ($true$ or $false$) of $A$ can be definitely determined. 
So, for point (1), $B\theta$ will be executed
with $C$ excluded; and for point (2), $C$ will be executed
with $B$ excluded. Therefore, the effect of {\em if-$A$-then-$B$-else-$C$}
is achieved. $\Box$

\begin{theorem}
\label{cut2}
The conditions of Prolog achieving the effect of 
if-$A$-then-$B$-else-$C$ using the following clauses
\begin{tabbing}
$\quad\quad$ \= $H \leftarrow A,!,B$.\\
\> $H \leftarrow C$.
\end{tabbing}
are the two conditions for TP-resolution plus a third one: (3) 
the evaluation of $A$ for its first answer will not go into a loop.
\end{theorem}

\noindent {\bf Proof.} Without loops in evaluating $A$ for its first answer,
the truth value ($true$ or $false$) of $A$ can be definitely determined. Otherwise,
neither $B$ nor $C$ will be executed, which violates the criterion of
Definition \ref{cut0}. $\Box$

By Theorem \ref{cut1}, for programs $P_{5_1}$, $P_{5_2}$ and $P_{5_3}$ (see
Example \ref{p5}) the two clauses $C_{np_1}$ and $C_{np_2}$ can be used 
by TP-resolution to represent {\em if-$p(X)$-then-$fail$-else-$true$}. 
By Theorem \ref{cut2}, however, Prolog cannot achieve such effect for $P_{5_3}$ because
the evaluation of $p(X)$ will go into a loop. Moreover, neither TP-resolution nor
Prolog can use $C_{p_1}$ and $C_{p_2}$ in $P_6$ (see Example \ref{p6}) to represent
{\em if-($q(X)$ and $p(b))$-then-$B$-else-$C$} because the evaluation of $p(b)$
will invoke $C$, which violates point (1) of Theorem \ref{cut1}.

Summarizing the above discussion leads to the following conclusion.

\begin{corollary}
Let $P$ be a program with the bounded-term-size property. 
If Prolog effectively handles cuts for $P$ 
w.r.t. the two intended purposes, so
does TP-resolution; but the converse is not true w.r.t. the first purpose.
\end{corollary}

\noindent {\bf Proof.} The second purpose of using cuts is achieved by
both Prolog and TP-resolution for any programs. For the first 
purpose, this corollary follows immediately 
from Theorems \ref{cut1} and \ref{cut2}. $\Box$

\section{Conclusions and Further Work}
Existing tabulated resolutions, such as OLDT-resolution, 
SLG-resolution and Tabulated SLS-resolution, rely on the 
solution-lookup mode in formulating tabling. 
Because lookup nodes are not allowed to resolve tabled subgoals against
program clauses, the underlying tabulated resolutions
cannot be linear, so that it is impossible to implement
such resolutions using a simple stack-based memory 
structure like that in Prolog. This may make their implementation
much more complicated (SLG-WAM for XSB is a typical example 
\cite{SSW98}, in contrast to WAM/ATOAM for
Prolog \cite{WAM83,ZHOU96}). Moreover, because lookup nodes totally depend
on solution nodes, without any autonomy, it may be difficult to handle
some strictly sequential operators such as cuts as effectively as
in Prolog (\cite{SSW94,SSWFR98}).

In contrast, TP-resolution presented 
in this paper has the following novel properties.
\begin{enumerate}
\item
It does not distinguish between solution and lookup nodes. Any
nodes can resolve tabled subgoals against program clauses as well
as answers in tables provided that they abide by the Table-first
policy, regardless of when and where they are 
generated.

\item
It makes linear tabulated derivations based on TP-strategy
in the same way as Prolog except that infinite loops are broken
and redundant computations are reduced. The resolution algorithm (Algorithm 2) 
is sound and complete for positive
logic programs with the bounded-term-size property and can be implemented
by an extension to any existing Prolog abstract
machines such as WAM \cite{WAM83} or ATOAM \cite{ZHOU96}.
\item
Due to its linearity, cuts can be easily realized. It handles cuts 
as effectively as Prolog in the case that cuts are used for pruning the search space,
and better than Prolog in the case for simulating the 
{\em if-then-else} statement.
\end{enumerate}

However, TP-resolution has some disadvantages. In particular,
an efficient implementation requires further investigation of the 
following issues. 
\begin{enumerate}
\item
Because it is a mixture of loop checking and tabling,
ancestor checking is required to see if a TP-derivation
has gone into a loop. That could be costly. Therefore, 
fast ancestor checking algorithms 
remain to be explored in further investigation.

\item
Answer iteration introduces redundant computations for those
programs and goals where the iteration is totally redundant (see, for
example, the programs $P_1$ and $P_2$ in Examples \ref{exp2} 
and \ref{p3} where 
no new answers can be derived through the iteration). 
Methods of determining in what
cases answer iteration can be ignored remain an interesting
open problem. 
\end{enumerate}

We have recently extended  
TP-resolution to compute the well-founded semantics of general
logic programs. A preliminary report on the extension appears in
\cite{shen99}. We are also working on the implementation of 
TP-resolution to realize a linear tabulated Prolog system.\\[.22in] 
{\Large \bf Acknowledgements}\\[.16in]
We are grateful to the three anonymous referees for their insightful 
comments, which have greatly improved the presentation.
The first author is supported in part by Chinese National
Natural Science Foundation and Trans-Century Training Program
Foundation for the Talents by the Chinese Ministry of Education.

\end{document}